%% file: aaai25.tex
\documentclass[letterpaper]{article} 
\usepackage{aaai25}  
\usepackage{times}  
\usepackage{helvet}  
\usepackage{courier}  
\usepackage[hyphens]{url}  
\usepackage{graphicx} 
\usepackage{natbib}  
\usepackage{caption} 
\usepackage{algorithm}
\usepackage{algorithmic}

\usepackage{newfloat}
\usepackage{listings}
\DeclareCaptionStyle{ruled}{labelfont=normalfont,labelsep=colon,strut=off} 
\floatstyle{ruled}
\newfloat{listing}{tb}{lst}{}
\floatname{listing}{Listing}
\usepackage[utf8]{inputenc} 
\usepackage[T1]{fontenc}    
\usepackage{hyperref}       
\usepackage{url}            
\usepackage{booktabs}       
\usepackage{amsfonts}       
\usepackage{nicefrac}       
\usepackage{microtype}      
\usepackage{xcolor}         

\usepackage{graphicx}
\usepackage{amsthm,amsmath,amssymb}
\usepackage{bm}
\usepackage{amsfonts}
\usepackage{algorithm}
\usepackage{algorithmic}
\usepackage{multirow}
\usepackage{tabularx}
\usepackage{subfigure}
\usepackage{wrapfig}
\usepackage{colortbl}
\usepackage[most]{tcolorbox}
\usepackage{listings}
\usepackage{xcolor}
\usepackage{pbox}
\usepackage{subcaption}
\usepackage{enumitem}
\usepackage{minitoc}
\usepackage{xcolor}
\usepackage{titletoc}
\usepackage{makecell}
\usepackage{tabularx}
\usepackage{setspace}

\definecolor{c1}{HTML}{0049C0}
\definecolor{c2}{HTML}{01FF2B}
\definecolor{c3}{HTML}{FF0000}
\definecolor{c4}{HTML}{548235}
\definecolor{mygray}{gray}{.95}
\newcommand{\ie}{\textit{i.e.}}

\DeclareMathOperator{\Encoder}{Encoder}
\DeclareMathOperator{\Center}{Center}
\DeclareMathOperator{\Kmeans}{K-means}
\DeclareMathOperator{\LLM}{LLM}

\newcommand{\symboldemo}{\raisebox{-0.2pt}{\includegraphics[scale=0.5]{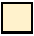}}}
\newcommand{\symboldemonwgm}{\raisebox{-0.2pt}{\includegraphics[scale=0.5]{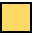}}}
\newcommand{\symbolquestion}{\raisebox{-0.2pt}{\includegraphics[scale=0.5]{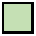}}}
\newcommand{\symbolcot}{\raisebox{-0.2pt}{\includegraphics[scale=0.5]{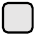}}}
\newcommand{\symbolcotcenter}{\raisebox{-0.2pt}{\includegraphics[scale=0.5]{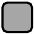}}}
\newcommand{\symbolcotimproved}{\raisebox{-0.2pt}{\includegraphics[scale=0.5]{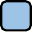}}}
\newcommand{\symbolanswer}{\raisebox{-0.2pt}{\includegraphics[scale=0.5]{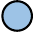}}}
\newcommand{\symbolanswerfinal}{\raisebox{-0.2pt}{\includegraphics[scale=0.5]{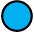}}}
\title{Causal Prompting: Debiasing Large Language Model Prompting based on Front-Door Adjustment}
\author{
    Congzhi Zhang\equalcontrib$^{\spadesuit}$\hspace{0.5mm}
    Linhai Zhang\equalcontrib$^{\heartsuit}$\hspace{0.5mm} 
    Jialong Wu\equalcontrib$^{\spadesuit}$\hspace{0.5mm}
    Yulan He$^{\heartsuit}$$^\clubsuit$\hspace{0.5mm}
    Deyu Zhou\thanks{~~Corresponding Author.}$^{\spadesuit}$\hspace{0.5mm} 
}
        
\affiliations{
    $^{\spadesuit}$\hspace{0.5mm}School of Computer Science and Engineering, Key Laboratory of Computer Network\\
    and Information Integration, Ministry of Education, Southeast University, China \\
    $^{\heartsuit}$\hspace{0.5mm} Department of Informatics, King’s College London, UK \\
    $^{\clubsuit}$\hspace{0.5mm} The Alan Turing Institute, UK \\
    \texttt{\{zhangcongzhi, jialongwu, d.zhou\}@seu.edu.cn} \\
    \texttt{\{linhai.zhang, yulan.he\}@kcl.ac.uk}
}

\begin{document}

\maketitle

\begin{abstract}
Despite the notable advancements of existing prompting methods, such as In-Context Learning and Chain-of-Thought for Large Language Models (LLMs), they still face challenges related to various biases. 
Traditional debiasing methods primarily focus on the model training stage, including approaches based on data augmentation and reweighting, yet they struggle with the complex biases inherent in LLMs.
To address such limitations, the causal relationship behind the prompting methods is uncovered using a structural causal model, and a novel causal prompting method based on front-door adjustment is proposed to effectively mitigate LLMs biases.
In specific, causal intervention is achieved by designing the prompts without accessing the parameters and logits of LLMs.
The chain-of-thought generated by LLM is employed as the mediator variable and the causal effect between 
input prompts and output answers is calculated through front-door adjustment to mitigate model biases.
Moreover, to accurately represent the chain-of-thoughts and estimate the causal effects, contrastive learning is used to fine-tune the encoder of chain-of-thought by aligning its space with that of the LLM. 
Experimental results show that the proposed causal prompting approach achieves excellent performance across seven natural language processing datasets on both open-source and closed-source LLMs.
\end{abstract}

\section{Introduction}
\label{sec:intro}

\begin{figure}
\includegraphics[width=0.48\textwidth]{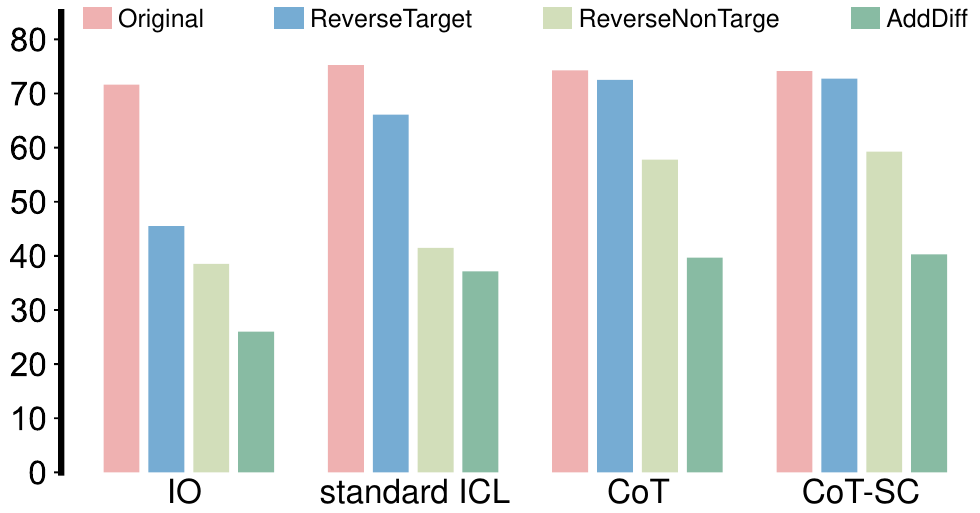}
\caption{Performance of different prompting methods on ABSA~\citep{pontiki2016semeval}
and its adversarial datasets on LLaMA-7b. ReverseTarget, ReverseNonTarget, and AddDiff denote three different adversarial transformations by TextFlint~\citep{wang2021textflint}. IO denotes the zero-shot setting where only the input question outputs the answer.} 
\vspace{-5mm}
\label{fig:introduction}
\end{figure}

\begin{figure*}[htbp]
\centering
\includegraphics[width=0.65\textwidth]{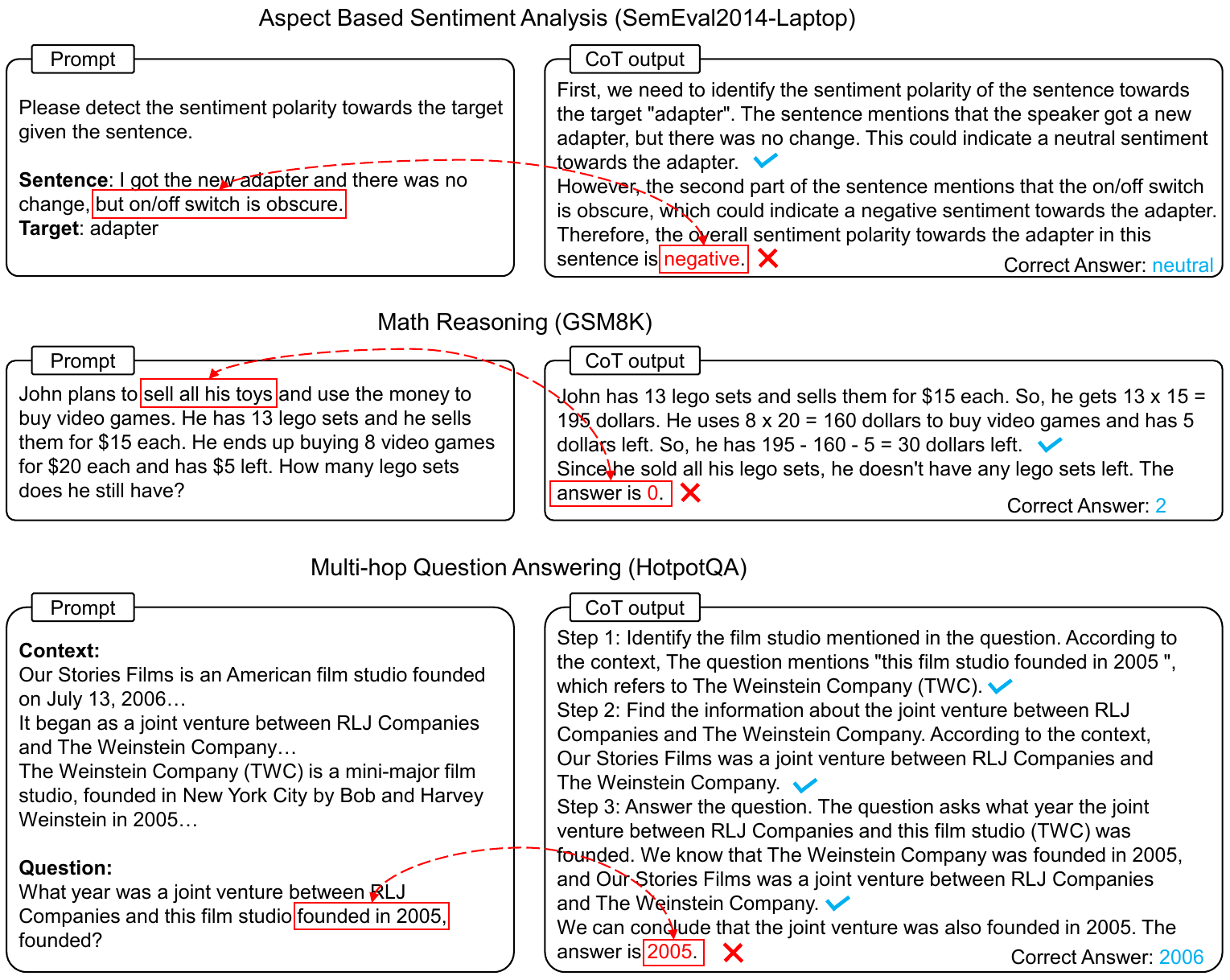}
\caption{LLMs suffer from bias in the pertaining corpus, leading them to rely on irrelevant text spans in prompts and generating incoherent chain-of-thoughts that harm the logical reasoning capability of the model. These examples were obtained by using the CoT prompting~\citep{wei2022chain} on the LLaMA3-8B model.} 
\label{fig:introduction_case}
\vspace{-5mm}
\end{figure*}

Large Language Models (LLMs) have shown remarkable emergent abilities, including In-Context Learning (ICL)~\citep{brown2020language,penglive,yang2024lever} and Chain-of-Thought (CoT) prompting~\citep{wei2022chain,wang2022self}, which allow LLMs to perform natural language tasks based on only a few instances without weight updating.
These prompting methods have achieved significant results across many traditional natural language processing tasks, including sentiment analysis, natural language inference, and machine reading comprehension~\citep{kojima2022large,zhou2022least,liu2023pre}. 

However, recent studies have shown that these advanced prompting methods are not robust enough~\citep{ye2023comprehensive} and can lead LLMs to produce hallucinatory results with incorrect or unfaithful intermediate reasoning steps~\citep{lyu-etal-2023-faithful,wang2023knowledge,bao2024llms,turpin2024language}.

Some studies~\citep{mallen-etal-2023-trust,wang2023resolving} believe that this phenomenon is due to the a conflict between the internal knowledge bias of LLMs and the external knowledge. 
Therefore, an effective solution is to interact with an external knowledge base to validate and adjust the reasoning process of LLMs~\citep{wang2023knowledge,zhang2023mitigating}.
Moreover, recent work debiases the chain-of-thoughts of LLMs by incorporating counterfactual knowledge and causal interventions~\citep{wu-etal-2024-decot}.
However, these methods are specifically tailored for knowledge-intensive tasks. Bias problems are also observed in other NLP tasks. 
As shown in Figure~\ref{fig:introduction_case}, in aspects-based sentiment analysis, mathematical reasoning, and multi-hop question-answering tasks, LLMs sometimes overly depend on certain text spans in the prompts, leading to wrong reasoning and answers. 
Notably, the first two tasks mentioned are not knowledge-intensive. We argue that LLMs fail to capture the true causal effect between questions and reasoning results and instead establish spurious correlations between certain text spans and answers.

In addition to the above qualitative analysis, our quantitative experiments also show that the current prompting methods are ineffective in addressing the bias issue.
As shown in Figure~\ref{fig:introduction}, the performance of all prompting methods drops significantly when evaluated on the corresponding adversarial dataset compared to the original dataset, indicating that LLMs may suffer from bias in the pertaining corpus. 
Moreover, it has been demonstrated that LLMs exhibit label bias, recency bias, and entity bias from context~\citep{zhao2021calibrate,wang-etal-2023-causal,fei2023mitigating}. 

Traditional debiasing methods mitigate the bias issue mainly during the model training stage, utilizing approaches such as data augmentation-based~\citep{Wei_Zou_2019,Lee_Won_Kim_Lee_Park_Jung_2021} and reweighting~\citep{schuster2019towards,Mahabadi_Belinkov_Henderson_2019}.
Data augmentation-based methods face challenges due to the cost and complexity of annotating bias cases, particularly limited by context length. 
Reweight-based methods encounter difficulties in assigning weights to each sample in prompt-based learning scenarios. 
Recently, debias methods based on causal inference~\citep{pearl2000models,pearl2022direct} have become popular because of their strict theoretical guarantees and good generalization. 
Causal inference-based methods only need to calibrate model prediction results during the inference stage~\citep{niu2021counterfactual,Tian_Cao_Zhang_Xing_2022,guo2022counterfactual,xu-etal-2023-counterfactual,chen2023causal}, which makes them well-suited for prompt-based learning scenarios.
However, counterfactual inference requires accessing LLM output logits, while back-door adjustment requires specific confounding variable values.

To address the aforementioned challenge, we propose to debias prompting methods through causal intervention using front-door adjustment~\citep{pearl2016causal}. 
Front-door adjustment enables causal intervention without the need to access confounding variable values or LLM output logits. 
As shown in Figure~\ref{fig:SCM}(a), the causal relationship behind the prompting method is uncovered using a structural causal model. 
Here $X$ denotes the input prompt, comprising demonstrations and test examples. 

$A$ denotes the predicted answer generated by the LLM. 
$U$ is the unobservable confounder that introduces various biases in the pertaining corpus. 

The debiasing process involves measuring the causal effect between the treatment $X$ and the outcome $A$. 
However, as $U$ absorbs complex biases of LLMs that are difficult to model or detect, back-door adjustment is not feasible for calculating the causal effect between $X$ and $A$. To address this issue, as shown in Figure~\ref{fig:SCM}(b), we use the chain-of-thought generated by LLM as the mediator variable $R$ between $X$ and $A$.

As Figure~\ref{fig:introduction_case} illustrates, while LLMs initially reason correctly, biases often confuse the final step of answer derivation.  
To simplify,  we ignore the edges between $U$ and $R$, aligning our causal graph with the front-door criterion~\citep{pearl2016causal}.
By this way, we can use the front-door adjustment to estimate the causal effect between $X$ and $A$ without accessing $U$.

Therefore, in this paper, we propose \textbf{Causal Prompting}, a novel prompting method for debiasing based on front-door adjustment.
Unlike previous causal inference-based methods, causal intervention is implemented by modifying prompts without accessing the parameters and logits of LLMs. 
Specifically, to estimate the causal effect between $X$ and $R$, we leverage self-consistency (SC)~\citep{wang2022self} of LLMs and a clustering algorithm to compute the probability of the chain-of-thought $R$. 
To measure the causal effect between $R$ and $A$, we use the normalized weighted geometric mean (NWGM) approximation~\citep{xu2015show} to select the optimal demonstration set, which can help the model to generate an unbiased answer. Overall, CoT, SC, and ICL are effectively combined through front-door adjustment to mitigate LLM biases in NLP tasks. 
Note that in the clustering and NWGM algorithms, an $\Encoder$ is needed to obtain the representations of chain-of-thoughts.
Since $\Encoder$ and LLMs have different semantic understanding of the chain-of-thought, we use contrastive learning~\citep{chen2020simple} to fine-tune the $\Encoder$ to align its representation space with LLMs to estimate causal effects more accurately.

The contributions of this work are summarized as follows:
\begin{itemize}[itemsep=0.5pt, parsep=0pt]
\vspace{-2mm}
\item 
Our work aims to identify and analyze the bias problem in LLM prompting methods from the perspective of causal inference, adhering more closely to the principles of the field. 
Moreover, the front-door adjustment is proposed to theoretically address the bias problem in prompting. 
\item Contrastive learning is proposed to fine-tune the $\Encoder$ of the chain-of-thoughts, aligning the space of the $\Encoder$ with LLMs to accurately capture representations of chain-of-thoughts and estimate causal effects. 
\item The proposed approach achieves excellent performance across seven natural language processing datasets using both open-source and closed-source LLMs.
\end{itemize}

\section{Preliminaries}
\label{sec:prel}

\subsection{Structural Causal Model and Causal Intervention}

A Structural Causal Model (SCM)~\citep{pearl2016causal} is used to describe the causal relationships between variables. In SCM, we typically use a directed acyclic graph $G = \{V, E\}$, where $V$ represents the set of variables and $E$ represents the set of direct causal relationships.

\begin{figure}
\includegraphics[width=0.48\textwidth]{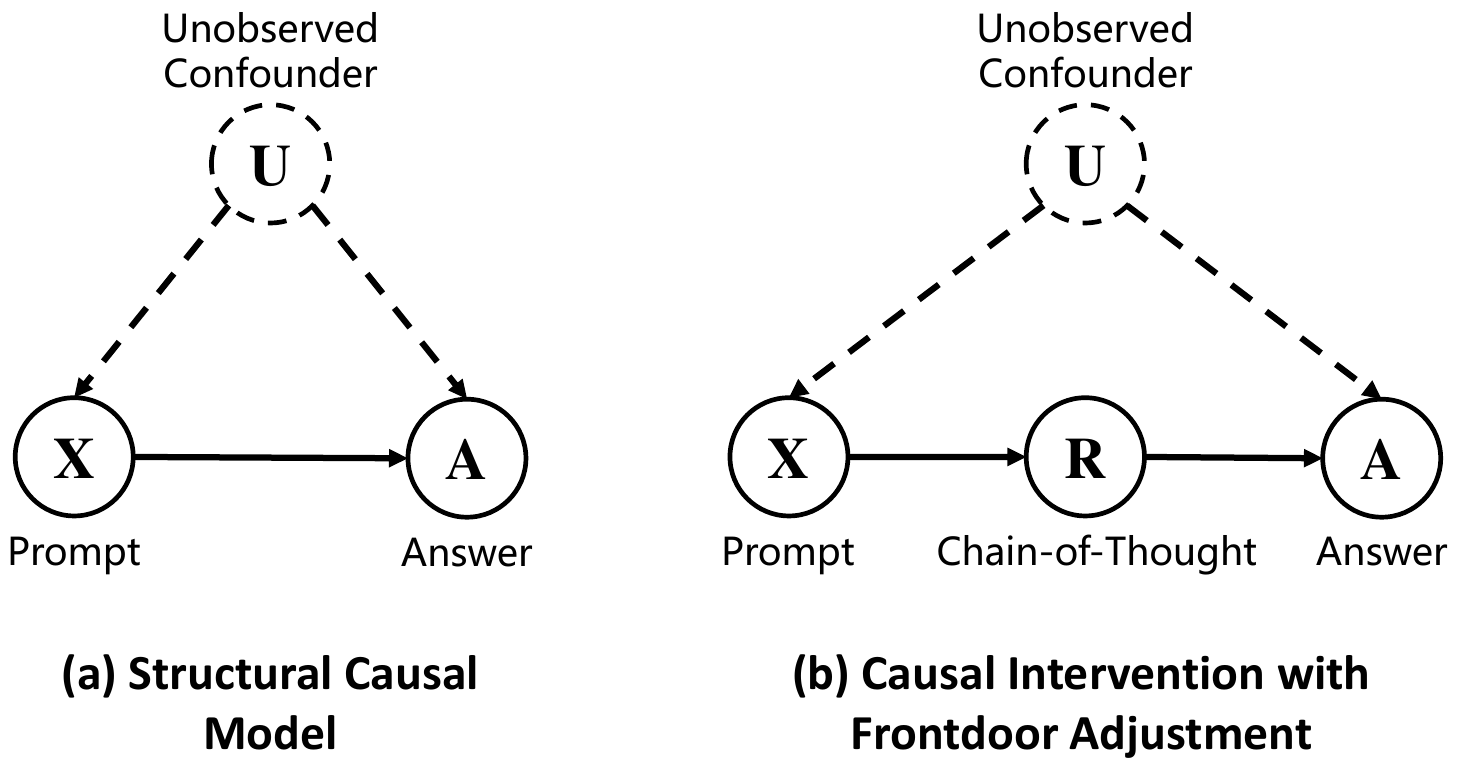}
\caption{Structural causal model for the prompting method.
(a) The causality of prompt and answer is confounded by unobservable variable.
(b) The chain-of-thought generated by LLMs as a mediator variable between prompt and answer.
}
\label{fig:SCM}
\vspace{-5mm}
\end{figure}

As shown in Figure~\ref{fig:SCM}(a), $X$ denotes the input prompt, including demonstrations and test examples. 
$A$ denotes the predicted answer generated by the LLMs. 
LLMs generate answers based on prompt, so we have $X \rightarrow A$, which means that $X$ is the direct cause of $A$.
LLMs might learn spurious 
correlations between text patterns and answers from pre-trained corpora or instruction-supervised fine-tuning datasets~\citep{xing2020tasty,li2024understanding,bao2024llms}, leading to bias in downstream tasks.
Previous work argues that the reason for this bias is that LLMs tend to follow a certain latent concept~\citep{xie2021explanation} or an implicit 
reasoning results~\citep{li2024understanding} in the reasoning process, rather than following the explicitly generated chain-of-thought. 
This leads to the final answer does not necessarily follow from the generated chain-of-thought, specifically, there is no actual causal relationship between the chain-of-thought and the answer~\citep{lyu-etal-2023-faithful,bao2024llms}.
To accurately calculate the causal effect between $X$ and $A$, we use the unobservable variable $U$ to describe this latent concept or implicit reasoning results, using the back-door path $X \leftarrow U \rightarrow A$ denotes that the causality of $X$ and $A$ is confounded by $U$.

In SCM, if we want to compute the true causal effect between two variables $X$ and $A$, we should block every back-door path between them~\citep{pearl2018book}. 
For example, as shown in Figure~\ref{fig:SCM}(a), we should block $X \leftarrow U \rightarrow A$ to obtain the true causal effect between $X$ and $A$.
We typically use causal interventions for this purpose, which use the $do$ operation to estimate the causal effect between $X$ and $A$. In the causal graph satisfying Figure~\ref{fig:SCM}(a), the $do$-operation can be computed by back-door adjustment~\citep{pearl2016causal}:
\begin{equation}
P(A|do(X)) = \sum_{u}P(A|X,u)P(u)
\label{back-door}
\end{equation}

\subsection{Front-door Adjustment}
Since confounding factor $U$ is inaccessible, back-door adjustment cannot be performed. Fortunately, the front-door adjustment~\citep{pearl2016causal} does not require access to the values of the confounding factor $U$ to calculate the causal effect between $X$ and $A$. 
As shown in Figure~\ref{fig:SCM}(b), we use the chain-of-thought generated by LLM as a mediator variable $R$ between $X$ and $A$.

In practice, as depicted in Figure~\ref{fig:introduction_case}, LLM can perform correct reasoning at the beginning, but it is often easily confused by bias in the last step of deriving the answer. Consequently, we decided to start with the simple SCM and focus on the confounder between $X$ and $A$. In order to simplify the causal graph, we ignore the confounder of $R$ with other variables, aligning our causal graph with the front-door criterion~\citep{pearl2016causal}.
According to the front door adjustment, $P(A|do(X))$ can be formulated as:
\begin{equation}
P(A|do(X)) = \sum_{r}P(A|do(r))P(r|do(X))
\label{eq:front-door-2parts}
\end{equation}
where $r \in R$ is the chain-of-thought generated by LLMs in response to the prompt $X$.
The causal effect between $X$ and $A$ is decomposed into two partially causal effects $P(r|do(X))$ and $P(A|do(r))$.

Next, we discuss how to estimate these two components separately.
The first component is $P(r|do(X))$, represents the probability distribution of the chain-of-thought $r$ given the intervention $do(X)$.
To compute $P(r|do(X))$, we need to block the backdoor path $X \leftarrow U \rightarrow A \leftarrow R$ between $X$ and $R$.
Since there exists a collision structure $U \rightarrow A \leftarrow R$, the backdoor path has been blocked~\citep{pearl2016causal} and we can get:
\begin{equation}
P(r|do(X)) = P(r|X)
\label{eq:front-door-part1}
\end{equation}
Now, we focus on the computation of the second component $P(A|do(r))$, represents the probability distribution of the answer $A$ given the intervention $do(r)$.
To compute $P(A|do(r))$, we need to block the backdoor path $R \leftarrow X \leftarrow U \rightarrow A$ between $R$ and $A$. Since we do not have access to the details of $U$, we implement back-door adjustments with the help of prompt $X$:
\begin{equation}
P(A|do(r)) = \sum_{x}P(x)P(A|r,x)
\label{eq:front-door-part2}
\end{equation}
where $x \in X$ denotes the input prompt, including demonstrations and test examples.

Finally, substituting Equations~\eqref{eq:front-door-part1} and \eqref{eq:front-door-part2} into Equation \eqref{eq:front-door-2parts} after we obtain the estimation of $P(r|do(X))$ and $P(A|do(r))$.
Hence, the final $P(A|do(X))$ can be represented as follows:
\begin{equation}
\begin{split}
    P(A|do(X))&=\sum_{r}{P(r|do(X))P(A|do(r))} \\
    &=\underbrace{\sum_{r}{P(r|X)}}_{CoT-SC}\underbrace{\sum_{x}{P(x)P(A|r,x)}}_{ICL}
\end{split}
\label{front_door}
\end{equation}
where the first component $\sum_{r}{P(r|do(X))}$ can be estimated by combining the CoT and SC prompting methods, and the second component $P(A|do(r))$ can be computed by selecting the demonstration examples in ICL prompting. 

\section{Method}
\label{sec:method}
\begin{figure*}[htbp]
 \centering
 \includegraphics[width=0.8\textwidth]{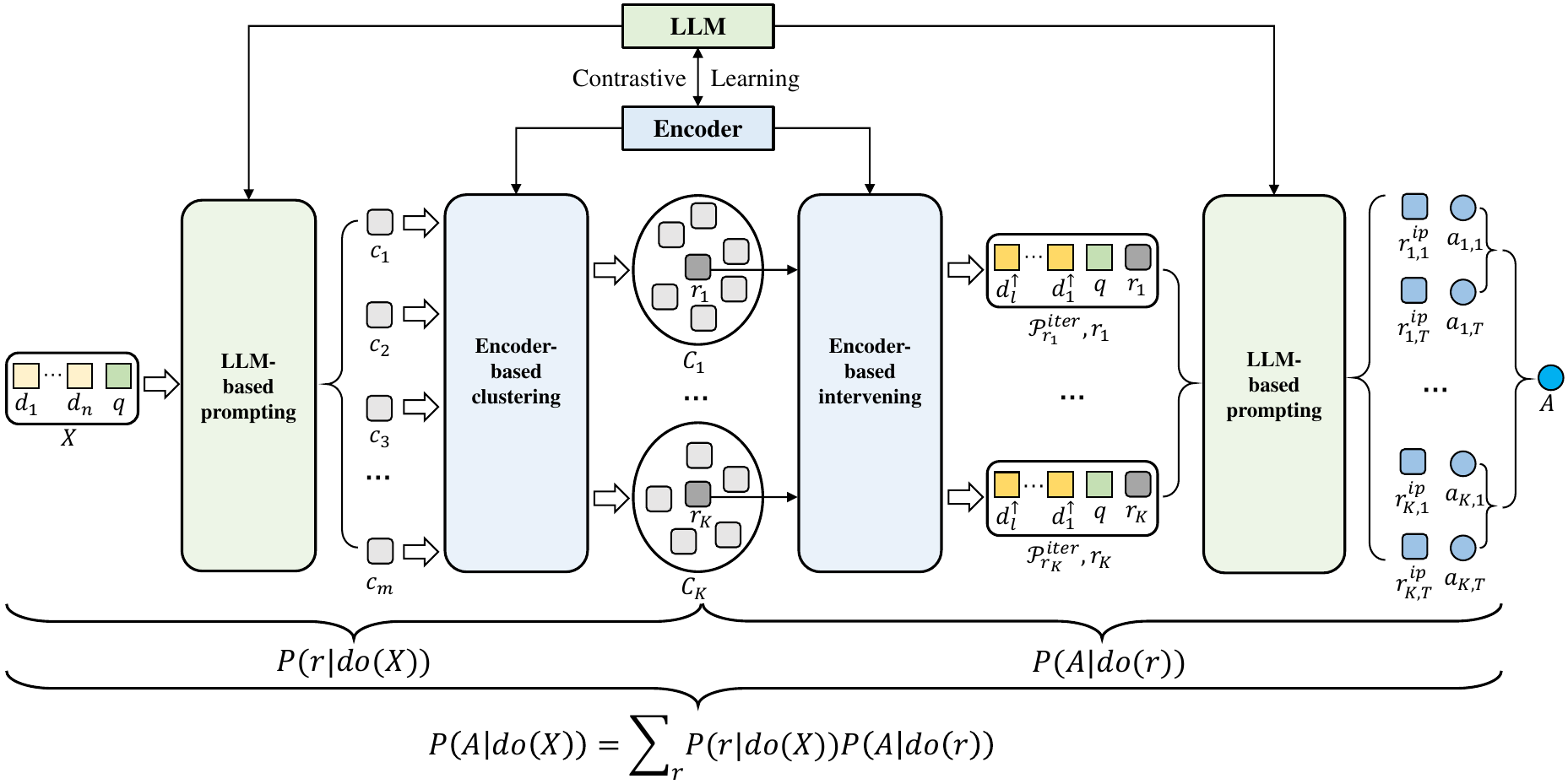}
 \caption{The overall framework of Causal Prompting.
Firstly, based on the input prompt $X$ consisting of the demonstration examples $\symboldemo$ and a question $\symbolquestion$ of the test example, we query the $\LLM$ to generate $m$ distinct CoTs $\symbolcot$. 
Then, these CoTs are clustered into $K$ clusters by an $\Encoder$-based clustering algorithm. 
Subsequently, $K$ representative CoTs $\symbolcotcenter$ are selected by searching the closest CoT to the cluster center.
Secondly, the optimal demonstration examples $\symboldemonwgm$ are retrieved for each representative CoT $\symbolcotcenter$ through the $\Encoder$-based intervention algorithm, and then the input prompt $\mathcal{P}_{r_k}^{iter}$ after the intervention is obtained.
Finally, we query the $\LLM$ $T$ times, obtaining $T$ improved CoTs $\symbolcotimproved$ and $T$ answers $\symbolanswer$ for each representative CoT $\symbolcotcenter$. The final answer $\symbolanswerfinal$ is obtained by performing a weighted voting.
}
\label{fig:framework}
\vspace{-3mm}
\end{figure*} 

As shown in Figure~\ref{fig:framework}, Causal Prompting aims to estimate the causal effect between input $X$ and answer $A$.
The estimation is achieved using the front-door adjustment, which divides the causal pathway into \textbf{two} distinct parts: the causal effect between $X$ and chain-of-thought $r$, and the causal effect between $r$ and $A$.

\textbf{First}, the causal effect between $X$ and chain-of-thought $r$, $P(r|do(X))$ is estimated by combining the Chain-of-Thought prompting with a $\Encoder$-based clustering algorithm.
\textbf{Second}, the causal effect between $r$ and $A$, $P(A|do(r))$ is estimated by combining the In-Context Learning prompting with the normalized weighted geometric mean (NWGM) approximation algorithm. 
The final answer is aggregated by performing a weighted voting algorithm.
Moreover, contrastive learning\citep{chen2020simple,gao2022improving,zhang2023multi} is employed to align the representation space of the $\Encoder$ and the LLMs for more precise estimation.

We will first introduce the estimation of $P(r|do(X))$ and $P(A|do(r))$, respectively, then combine them to derive $P(A|do(X))$. Finally, we will discuss how we align the representation space between the $\Encoder$ and the $\LLM$.

\subsection{Estimation of $P(r|do(X))$}
\label{sec:p1}
We firstly undertake the estimation of $P(r|do(X))$.
$P(r|do(X))$ measures the causal effect between input $X$ and chain-of-thought $r$. 
As shown in Equation~\eqref{eq:front-door-part1}, the estimation of $P(r|do(X))$ is equivalent to the estimation of $P(r|X)$.
However, $P(r|X)$ is still intractable for LLMs.
On the one hand, the output probability is often inaccessible for most closed-source LLMs; on the other hand, the chain-of-thoughts $r$ are challenging to enumerate comprehensively.
Therefore, to estimate the causal effect $P(r|do(X))$ for both open-source and closed-source LLMs, we employ the CoT prompting and integrate it with a clustering algorithm.
To be more specific, we initially prompt the LLMs to generate multiple CoTs based on the input.
The prompts for CoTs generation are detailed in Appendix~\ref{sec:prompt_cot}.
Subsequently, the CoTs are projected into embeddings.
The embeddings are then clustered to form distinct groups based on their similarity.
Finally, the centroid of each cluster is selected as the optimal and representative chain-of-thought.
The probability associated with each representative chain-of-thought is then estimated based on the size of its respective cluster.

To enhance the quality of generated CoTs, $n$ in-context demonstrations $d$ are selected from training set based on question similarity.
These demonstrations are then concatenated with the test question $q^{test}$ to form the final prompt.
Thus, the final prompt $\mathcal P$ is structured as follows:
\begin{equation}
   \mathcal P = [d_{1},...,d_{n},q^{test}]
    \label{eq:input_prompt}
\end{equation}
where each $d_i=(q^{demo}_i,r^{demo}_i)$ contain the  demonstration question $q^{demo}_i$ and its corresponding demonstration chain-of-thought $r^{demo}_i$. Where $i \in \{1,...,n\}$, $n$ denotes the number of demonstration examples in few-shot prompt method. 
In the practical implementation, we use prompt $\mathcal{P}$, which is fed into the LLMs to represent $X$ in the structural causal model.

Based on the input prompt $\mathcal P$, LLMs are prompted to generate $m$ distinct CoTs $c$ by increasing the temperature parameter of LLMs.
This adjustment encourages more diverse outputs, where the same procedure is also employed in self-consistency prompting of LLMs~\cite{wang2023selfconsistency}.
In this way, we can obtain the set of chain-of-thoughts as follows:
\begin{equation}
    \{c_i|i=1,...,m\} = \LLM(\mathcal{P})
\end{equation}
To perform the distance-based clustering method, the generated CoT $c_i$ are further fed into a $\Encoder$ to get the text embedding $\overline{c}_i$.
Following the previous work~\citep{devlin2018bert}, the input is concatenated with the special tokens \small [CLS] and \small [SEP], and the embedding of the \small [CLS] token is taken as the embedding of CoT $c_i$.
\begin{equation}
    \overline{c}_i = \Encoder(\text{\small [CLS]},c_i,\text{\small [SEP]})
\end{equation}
Then K-means clustering algorithm~\citep{har2005smaller,wu2023self} is applied to the embeddings to get $K$ clusters $C$ as follows:
\begin{equation}
    \{C_1,...,C_K\} = \Kmeans(\overline{c}_1,...,\overline{c}_m)
\end{equation}
where $C_k$ refers to the $k$-th cluster of the clustering result, $K$ denotes the number of clusters.

Based on the clusters, $K$ representative chain-of-thoughts $r$ are selected by searching the closest chain-of-thought to the cluster center. 
\begin{equation}
    r_k = \Center(C_k), k=1,...,K
    \label{eq:select_cots}
\end{equation}
The causal effect between input $X$ and chain-of-thought $r_k$ is estimated based on the cluster size as follows:
\begin{equation}
    P(r_k|do(X)) \approx \frac{|C_k|}{m}
    \label{eq:do1}
\end{equation}
where $|C_k|$ denotes the size of cluster $C_k$. 

\subsection{Estimation of $P(A|do(r))$}
\vspace{-1mm}
\label{sec:p2}
Based on the $K$ chain-of-thoughts selected by Equation~\eqref{eq:select_cots} in Section~\ref{sec:p1}, we estimate $P(A|do(r_k))$ for each chain-of-thought $r_k$. 
For convenience, we omit the subscript $k$ and use $P(A|do(r))$ to denote $P(A|do(r_k))$ in the following.
$P(A|do(r))$ measures the causal effect between the chain-of-thought $r$ and the answer $A$.
Based on the discussion in Equation~\eqref{eq:front-door-part2}, $P(A|do(r))$ can be calculated with backdoor adjustment as follows:
\begin{equation}
\begin{split}
    P(A|do(r)) = \sum_{x\in X}{P(x)P(A|r,x)} 
    = \mathbb{E}_{x \in X}[P(A|r,x)]
\end{split}
\label{eq:12}
\end{equation}
where $P(A|r,x)$ denotes the probability of the final answer $A$ generated by LLM based on the given prompt $x$ and the chain-of-thought $r$. 

However, the value space of $X$ is inexhaustible in most of the cases, and previous work employs the normalized weighted geometric mean (NWGM) approximation~\citep{xu2015show,Tian_Cao_Zhang_Xing_2022,chen2023causal} to tackle this problem, where a confounder embedding $\overline{x}^{'}$ is estimated to approximate the expectation of variable $X$.
\begin{equation}
\begin{split}
    \mathbb{E}_{x\in X}[P(A|r,x)]  \approx P(A|r,\mathbb{E}_{x\in X}[x]) 
     \approx P(A|concat(r,\overline{x}^{'}))
\end{split}
\end{equation}
where $concat(\cdot,\cdot)$ denotes vector concatenation, $\overline{x}^{'}$ denotes the confounder embedding of $X$.

Inspired by the previous works~\citep{xu2015show,Tian_Cao_Zhang_Xing_2022,chen2023causal,zhang2024causal}, we propose a prompting version of NWGM approximation to perform the back-door adjustment for LLMs prompting by combining a $\Encoder$-based intervention and In-Context Learning (ICL) prompting.
The original idea of NWGM is to augment the representation of the chain-of-thought $r$ with an embedding $\overline{x}^{'}$ that contains all sample information as much as possible. 
However, at the prompting level, we cannot include all samples in context due to the limited context length, so we use only those samples that are most useful for improving the current chain-of-thought $r$.

Specifically, we use the $\Encoder$ to obtain the embedding $\overline{r}_{k}$ of the $k$-th chain-of-thought $r_k$.
Subsequently, ICL demonstrations are selected by searching the entire training set based on the chain-of-thought embedding $\overline{r}_{k}$ to approximate the effect of taking expectations on input $X$. 
Finally, we rank the ICL demonstrations according to their similarity weights to indicate the importance of different samples.

Note that, as shown in Equation~\eqref{eq:input_prompt}, the input prompt $\mathcal{P}$ includes demonstrations $d$ and test question $q^{test}$. 
Directly modifying the certain text span of test examples will change the semantics of question $q^{test}$. Therefore, we only modify the demonstrations $d$ and implement the NWGM approximation by In-Context Learning.
In fact, the goal of our prompting version of the NWGM algorithm is to enable the LLMs to learn from the demonstrations how to improve the chain-of-thought $r$ of the test example.
As shown in the prompt template in Appendix~\ref{sec:prompt_causal_prompt}, we introduce both wrong and correct chain-of-thoughts of demonstrations.

Given a training set $\mathcal D=\{d_j=(q_j,r_j^{wrong},r_j^{correct})\}_{j=1}^{N}$, and a chain-of-thought $r_k$ of test example, where $q_j$ denotes the question of $j$-th training sample, $r_j^{wrong}$ and $r_j^{correct}$ denote the wrong and correct chain-of-thoughts of demonstration $d_j$, $N$ denotes the size of the training set, $r_k$ refers to the $k$-th chain-of-thought selected by Equation~\eqref{eq:select_cots} in Section~\ref{sec:p1}.
The embedding $\overline{r}_k$ of chain-of-thought $r_k$ and the embedding $\overline{d}_j$ of demonstration $d_j$ are obtained by the following:
\begin{equation}
\begin{split}
    \overline{r}_k &= \Encoder(\text{\small [CLS]}, r_k,\text{\small [SEP]}) \\
    \overline{d}_j &= \Encoder(\text{\small [CLS]},r_j^{wrong},\text{[\small SEP]}) \\
\end{split}
\end{equation}

Previous works~\citep{margatina2023active,liu-etal-2022-makes} have shown that using demonstration examples that are semantically similar to the test examples allows better performance for In-Context Learning. Therefore, the back-door intervention is approximated by searching the most similar instance based on chain-of-thought embedding $\overline{r}_k$.
Specifically, we sort the training set $\mathcal D$ from largest to smallest according to the cosine similarity between $\overline{r}_k$and $\overline{d}_j$.
\begin{equation}
\begin{split}
    \{d_j^{\uparrow}\}_{j=1}^N &= Sort(\mathcal D, \overline{r}_k, \{\overline{d}_j\}_{j=1}^{N}) \\
\end{split}    
\end{equation}
where $d_j^{\uparrow}$ denotes the sorted demonstration example, $Sort$ means that, given a predefined cosine similarity function $cos$, the samples are ordered so that $cos(\overline{r}_k, \overline{d}_i) \ge cos(\overline{r}_k, \overline{d}_j)$ when $i < j$.

Then the $l$ most similar demonstration examples are selected to concatenate into prompt, where $l\ll N$.
Note that, unlike the KATE~\citep{liu2021makes} method, we put the most similar demonstration samples closer to the test samples because this order is more beneficial for our NWGM algorithm to learn information for improving the chain-of-thoughts from the demonstration based on practical experiments, detailed in Appendix~\ref{sec:ablation}.
For each chain-of-thought $r_k$ of a test sample, the final input prompt after intervention is given as follows:
\begin{equation}
    \mathcal{P}_{r_k}^{iter} = [d_l^{\uparrow},...,d_{1}^{\uparrow},q^{test}]
    \label{eq:x_it}
\end{equation}

Subsequently, we query the LLMs $T$ times, obtaining $T$ answers and $T$ improved chain-of-thoughts using the prompt $\mathcal{P}_{r_k}^{iter}$ and chain-of-thought $r_k$.
\begin{equation}
    \{(r_{k,t}^{ip}, a_{k,t})| t=1,...,T\} = \LLM(\mathcal{P}_{r_k}^{iter},r_k)
\end{equation}
where $r_{k,t}^{ip}$ denotes the $t$-th improved chain-of-thought for chain-of-thought $r_k$.

We then use majority voting to estimate the probability of the answer as follows:
\begin{equation}
    P(A|do(r_k)) \approx \frac{\sum_{t=1}^T \mathbb{I}(A=a_{k,t})}{T}
    \label{eq:do2}
\end{equation}

\subsection{Estimation of $P(A|do(X))$}
Based on the results of Equation~\eqref{eq:do1} in Section~\ref{sec:p1} and Equation~\eqref{eq:do2} in Section~\ref{sec:p2}, the final answer is obtained by performing a weighted voting as follows:
\begin{equation}
\begin{split}
    P(A|do(X)) &= \sum_{r_k} P(r_k|do(X))P(A|do(r_k))  \\
    &=  \sum_{k=1}^K \frac{|C_k|}{m} \cdot \frac{\sum_{t=1}^T \mathbb{I}(A=a_{k,t})}{T}
\end{split}
\end{equation}
Finally, we chose the answer with the largest weight as the final answer.
In this way, with the front-door adjustment, we calibrate the probability distribution $P(A|X)$ obtained by the CoT-SC method to $P(A|do(X))$ obtained by the Causal Prompting method.
Algorithm 1 in Appendix~\ref{sec:algo} shows the overall prompting process.
Cases in Appendix~\ref{sec:case_study} show the overall flow and intermediate step output of Causal Prompting on mathematical reasoning and multi-hop question answering datasets.

\subsection{Representation Space Alignment}
\label{sec:alignment}
In the clustering discussed in Section~\ref{sec:p1} and NWGM algorithm presented in Section~\ref{sec:p2}, an $\Encoder$ is needed to derive the representations of chain-of-thoughts. However, the semantic representation of $\Encoder$ and $\LLM$ differ significantly. 
Two chain-of-thoughts that $LLM$ considers similar may not be close in the representation space of the $\Encoder$.
As illustrated in Figure~\ref{fig:cluster_visualization} in Appendix~\ref{sec:dis}, the chain-of-thoughts generated by $\LLM$ are not distinctly separable in the representation space of the vanilla $\Encoder$.

To align the representation spaces of the $\Encoder$ and the LLMs, we take each chain-of-thought $r$ in the training dataset $\mathcal{D}$ as an anchor, use LLM to generate the corresponding positive samples, use the other samples within the batch as negative samples, and then use contrastive learning to fine-tune the $\Encoder$. 
The prompt template used to generate positive samples is detailed in Appendix~\ref{sec:prompt_contrastive_learning}.

For chain-of-thought $r$, we prompt the LLM to generate a similar sentence $r^{+}$ as the positive sample.
Following previous works~\citep{gao2022improving,zhang2023multi}, we use the InfoNCE loss~\citep{chen2020simple} to fine-tune the $\Encoder$ :
\begin{equation}
    \sum_{\overline{r}_p \in Pos(r)}{-log\frac{g(\overline{r},\overline{r}_p)}{g(\overline{r},\overline{r}_p)+\sum_{j\in Neg(r)}{g(\overline{r},\overline{r}_j)}}}
\end{equation}
where the $\overline{r}$ and $\overline{r}_p$ are the representations of $r$ and its positive samples. 
$Pos(r)$ and $Neg(r)$ refer to the positive set and the negative set for the chain-of-thought $r$.
$Pos(r)=\{\overline{r}_{p1},\overline{r}_{p2}\}$, where $\overline{r}_{p1}$ is augmented representation of the same chain-of-thought $r$, obtained with different dropout masks, and $\overline{r}_{p2}$ is the representation of positive sample $r^{+}$. $j \in Neg(r)$ is the index of in-batch negative samples. $g$ is a function: $g(\overline{r},\overline{r}_p) = exp(\overline{r}^{T}\overline{r}_p/temp)$, where $temp$ is a positive value of temperature in the contrastive learning.

\section{Experiments}

\begin{table*}[]
\centering
\small

\begin{tabular}{@{}l|cc|cccc|ccc@{}}
\toprule
 &
  GSM8K &
  MATH &
  \multicolumn{2}{c}{HotpotQA} &
  \multicolumn{2}{c|}{MuSiQue} &
  ABSA &
  NLI &
  FV \\ \midrule
 Method & Acc   & Acc   & EM    & F1    & EM    & F1    & Acc   & Acc   & Acc   \\ \midrule
\rowcolor{mygray} \multicolumn{10}{c}{LLaMA2} \\
\midrule
Standard ICL & 6.14  & 3.71  & 41.20  & 59.56 & 26.09 & 41.16 & 47.26 & 28.20  & 56.87 \\
CoT & 27.07 & 4.72  & 44.70  & 64.84 & 18.71 & 30.27 & 49.12 & 27.56 & 70.07 \\
CoT-SC       & 31.92 & 6.32  & 49.30  & 68.53 & 31.16 & 46.36 & 53.70 & 33.57 & 72.20  \\
Causal Prompting & \textbf{36.47} & \textbf{8.76} & \textbf{52.20} & \textbf{70.88} & \textbf{34.68} & \textbf{48.79} & \textbf{67.55} & \textbf{50.83} & \textbf{81.07} \\ \midrule \rowcolor{mygray} \multicolumn{10}{c}{LLaMA3} \\
\midrule
Standard ICL & 18.65 & 14.24 & 37.20  & 62.17 & 17.42 & 24.22 & 72.14 & 63.75 & 80.67 \\
CoT          & 74.07 & 40.35 & 48.90  & 72.75 & 38.88 & 54.38 & 71.55 & 64.19 & 81.80  \\
CoT-SC       & 82.41 & 56.61 & 52.70  & 75.43 & 41.37 & 59.78  & 75.92 & 65.15 & 83.87 \\
Causal Prompting & \textbf{87.95} & \textbf{62.76} & \textbf{58.50} & \textbf{78.18} & \textbf{48.07} & \textbf{64.23} & \textbf{79.06} & \textbf{67.97} & \textbf{86.67} \\ \midrule
\rowcolor{mygray} \multicolumn{10}{c}{GPT-3.5} \\
\midrule
Standard ICL & 33.74 & 23.08 & 2.10   & 3.68  & 28.84 & 39.27 & 69.26 & 53.52 & 75.33 \\
CoT          &  71.87 & 53.50 & 11.70  & 16.49 & 41.37 & 57.82 & 65.74 & 63.55 & 80.67 \\
CoT-SC       & 80.21 & 58.38 & 41.60  & 56.82 & 46.27 & 60.83 & 74.59 & 66.88 & 82.73 \\
Causal Prompting & \textbf{85.44} & \textbf{70.18} & \textbf{58.20} & \textbf{78.10} & \textbf{50.13} & \textbf{65.40} & \textbf{80.13} & \textbf{71.93} & \textbf{86.53} \\ \bottomrule
\end{tabular}
\caption{The comparison results of Causal Prompting against baselines across different backbone LLMs, including LLaMA2, LLaMA3 and GPT-3.5, on seven datasets. 
The best results are in bold.}
\label{tab:results-main}
\end{table*}

\subsection{Datasets}

We evaluate the effectiveness of our approach on three tasks: \textbf{Math Reasoning} (GSM8K~\citep{cobbe2021training}, MATH~\citep{hendrycksmath2021}), \textbf{Multi-hop Question Answering} (HotpotQA~\citep{yang-etal-2018-hotpotqa}, MuSiQue~\citep{trivedi2021musique}), and \textbf{Natural Language Understanding} (Aspect-based Sentiment Analysis (ABSA)~\citep{pontiki2016semeval}, Natural Language Inference (NLI)~\citep{williams2017broad}, and Fact Verification (FV)~\citep{thorne2018fever}). 
For the NLU tasks, we use the original datasets (in-distribution, ID) and the corresponding adversarial datasets (out-of-distribution, OOD) to verify the robustness of our method.
Further details regarding the datasets are provided in Appendix~\ref{sec:dataset}.
The details regarding the evaluation can be found in Appendix~\ref{sec:eval}.

\subsection{Baselines}
We compare our approach with three other few-shot prompting approaches to evaluate its effectiveness:
Standard ICL, CoT and CoT-SC.
Their detailed settings are presented in Appendix~\ref{sec:baseline}.
Detailed settings and implementations of our method \textbf{Causal Prompting} can be found in Appendix~\ref{sec:settings} and Appendix~\ref{sec:implementation}.

\subsection{Main Results}
Table~\ref{tab:results-main} shows the comparison results between causal prompting and the aforementioned baselines.
Expectedly, the performance of Standard ICL, CoT, and CoT-SC improves progressively, as each subsequent method is an enhanced version of its predecessor. 
It not only confirms the effectiveness of integrating CoT into ICL, consistent with~\citep{brown2020language,wei2022chain,zhou2022least}, but also validates the efficacy of employing multiple sampling and voting strategies~\citep{wang2022self}.
\textbf{Causal Prompting} consistently delivers the best results across all metrics and datasets.
It indicates that our prompting method can comprehensively improve the ability of LLM in all three tasks.
Specifically, our method exhibits a more pronounced improvement in Math Reasoning and Multi-hop Question Answering tasks, with an average performance enhancement of approximately \textbf{5\%-10\%}. 
This substantial increase underscores our method's greater efficacy in tackling more challenging problems.

\subsection{Robustness Study}
\label{sec:robustness}
\begin{table}[]
\small
\centering
\begin{tabular}{@{}l|cc|cc|cc@{}}
\toprule
             & \multicolumn{2}{c|}{ABSA}        & \multicolumn{2}{c|}{NLI} & \multicolumn{2}{c}{FV} \\
             \midrule
Methods      & \texttt{Ori}            & \texttt{Adv}  & \texttt{Ori}    & \texttt{Adv}   &\texttt{Ori}  & \texttt{Adv}   \\ \midrule
ICL & 75.71          & 70.30   & 76.30   & 50.27  & 90.00  & 76.00  \\
CoT          & 77.27          & 68.60  & 74.81  & 54.77  & 91.40  & 77.00   \\
CoT-SC       & \textbf{80.56} & 73.53   & 76.17  & 57.16  & 93.40  & 79.10  \\ \midrule
Ours & 79.78 & \textbf{78.69} & \textbf{76.67} & \textbf{58.62}  & \textbf{95.40} & \textbf{82.30}  \\ \bottomrule
\end{tabular}
\caption{The results of the robustness study on LLaMA3. \texttt{Ori} denotes the original dataset (ID) and \texttt{Adv} denotes the adversarial dataset (OOD). The best results are in bold.}
\label{tab:results-robustness}
\end{table}

Recent causal-based works~\cite{Tian_Cao_Zhang_Xing_2022,zhang2024causal,zhu2023causal,wang-etal-2023-causal,xu-etal-2023-counterfactual,niu2021counterfactual,schuster2019towards,wu2024diner} have shown that using symmetric and adversarial (out-of-distribution) datasets \textbf{can evaluate the debiasing ability of models}.
Following their practice, we evaluate \textbf{Causal Prompting} on both original data and adversarial data of the NLU tasks, respectively.
Tables~\ref{tab:results-robustness} show the performance comparison results of our method and baselines on LLaMA3 model.
Although the performance of Causal Prompting decreases on \texttt{Ori} of ABSA, the improvement is larger on \texttt{Adv} data, resulting in the highest overall performance, see in~\ref{tab:results-main}.
This phenomenon aligns with findings reported in previous work on causal inference~\citep{Tian_Cao_Zhang_Xing_2022,wang-etal-2023-causal}.
It can be observed that the \texttt{Adv} of Causal Prompting is the highest on all datasets. 
This shows that our method generalizes well for both synthetic adversarial data in ABSA and NLI generated by TextFlint~\citep{wang2021textflint} and human-annotated real adversarial data in FV.
This further validates the robustness of our model in handling datasets with significant bias.

\subsection{More Experimental Results}

For details on robustness studies, ablation experiments and hyperparameter experiments, please refer to the appendix:
\begin{itemize}
    \item In Appendix~\ref{sec:more_robustness}, we report the robustness study results of LLaMA2 and GPT3.5.
    \item In Appendix~\ref{sec:ablation}, we perform a detailed ablation analysis to evaluate three pivotal aspects: (1) the effectiveness of the NWGM approximation, (2) the impact of incorporating contrastive learning, (3) the impact of K-means clustering and weighting mechanism.
    \item In Appendix~\ref{sec:hyper_study}, we conduct additional hyperparameter experiments to explore the impact of the number of clusters $K$ and the number of CoTs $m$ on the performance.
\end{itemize}

\subsection{Discussion}

In Appendix~\ref{sec:related_works}, we discuss related works on the topics of prompting strategies and debiasing with causal inference.
In Appendix~\ref{sec:dis}, we delve into five critical discussion questions (DQs) that are essential for understanding the contributions and limitations of our approach: (\textbf{DQ1}) In-depth analyses regarding the optimization of computational costs. (\textbf{DQ2}) The impact of performance threshold adjustments. (\textbf{DQ3}) The effects of contrastive learning. (\textbf{DQ4}) The rationale behind baseline selection. (\textbf{DQ5}) The new bias from the CoT generated by LLMs is not considered.

\section{Conclusion and Future Work}
We introduced Causal Prompting, a novel method for debiasing LLMs in NLP tasks by utilizing front-door adjustment in this work.
The CoT generated by LLMs is employed as a mediator variable in the causal graph. 
Specifically, the causal effect between input prompt and output answer is decomposed into two distinct components, the causal effect from the input prompt to CoTs and from CoTs to the answer.
The former component is estimated by combining the CoT prompting with a $\Encoder$-based clustering algorithm. 
The latter component is estimated by combining the ICL prompting with the NWGM approximation algorithm.
Moreover, Contrastive learning is used to fine-tune the $\Encoder$ so that the representation space of the $\Encoder$ is aligned with the LLM to estimate the causal effect more accurately. 
Our experimental results demonstrate that Causal Prompting significantly improves performance across seven NLP tasks on both open-source and closed-source LLMs.

Causal Prompting addresses inherent biases in LLMs by scaling up during the inference phase, building upon the theory of front-door adjustment.
This approach, which both enhances performance and yields debiased responses, aligns with the trend of obtaining optimal results at test time~\cite{snell2024scaling,o1,qu2024recursive,lightman2023let}.
It can be extended to a broader range of scenarios, such as safety or alignment, under theoretical guidance.

\section*{Acknowledgments}
The authors would like to thank the anonymous reviewers for their insightful comments. This work is funded by the National Natural Science Foundation of China (62176053). This work was supported in part by the UK Engineering and Physical Sciences Research Council (EPSRC) through a Turing AI Fellowship (grant no. EP/V020579/1, EP/V020579/2) and Innovate UK through the Accelerating Trustworthy AI programme (grant no. 10093055). This work is supported by the Big Data Computing Center of Southeast University.

\bibliography{aaai25}


\input{appendix.tex}

\end{document}

%% file: appendix.tex
\clearpage 
\onecolumn 
\appendix

\section*{\centering \Large Technical Appendix}

\startcontents[appendices]
\section*{\centering \Large Contents}
\begin{spacing}{1.5} 
\printcontents[appendices]{l}{1}{\setcounter{tocdepth}{2}}
\end{spacing}
\newpage
\clearpage

\section{Discussion}
\label{sec:dis}
\begin{figure}[t]
\centering
\subfigure[GSM8K]{\includegraphics[width=0.4\textwidth]{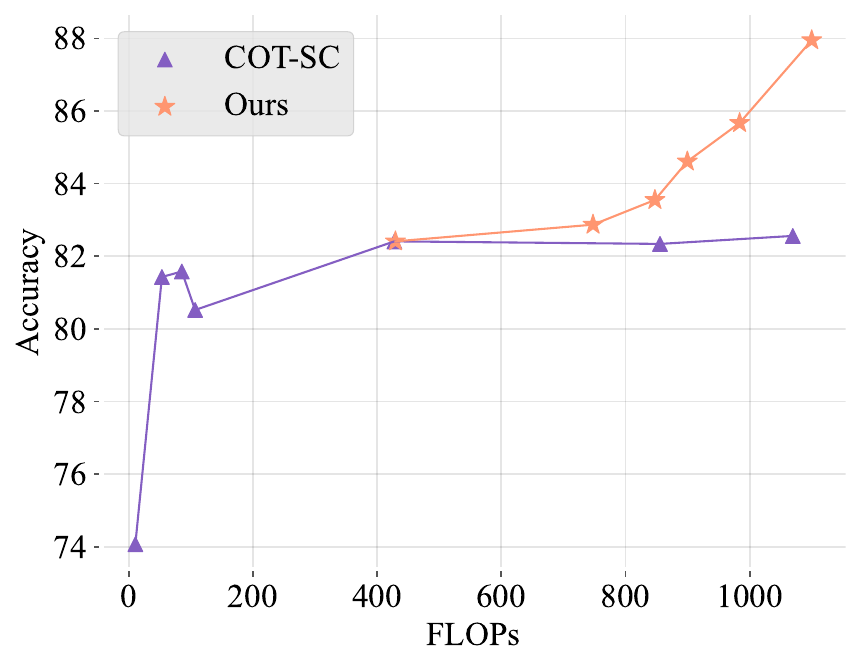}}
\subfigure[MuSiQue]{\includegraphics[width=0.4\textwidth]{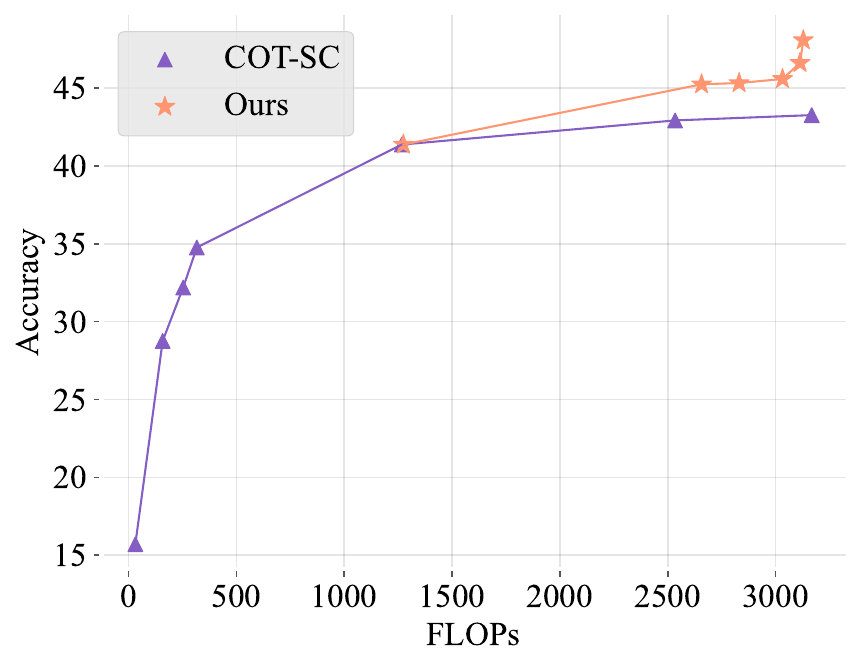}}
\caption{Comparison of FLOPs cost between Causal Prompting and CoT-SC method on LLaMA3. }
\label{fig:cost}
\end{figure}
In this section, we will address the following Discussion Questions (\textbf{DQ}) to elucidate our contributions more clearly.

\paragraph{DQ1: How can we reduce computational costs incurred by additional operations?\\}
We acknowledge that our proposed method requires more
cost than CoT-SC or direct prompting. 
Compared to CoT-SC, Causal Prompting involves numerous additional operations, resulting in increased cost overhead.
However, our approach is relatively cost-effective and the improved performance potentially leads to more faithful outcomes.
Importantly, \textbf{not} every test instance necessitates a front-door adjustment.
To determine whether a a front-door adjustment is needed for a given example, we employ a \textit{self-consistency metric}, defined as the proportion of answers receiving the majority of votes.
A front-door adjustment is executed if the \textit{self-consistency metric} falls below a predefined threshold $s$.
As shown in Figure~\ref{fig:cost}, adjusting the threshold $s$ allows us to control the cost of the entire test dataset.
In our experiments, the computational cost is quantified in terms of Floating Point Operations per Second (FLOPs). 
For a test set containing $1000$ samples with an average sample length of $500$, the inference time on LLaMA3 is approximately two hours using the vLLM framework~\citep{kwon2023efficient}.
All computations using open-source LLMs were executed on an NVIDIA A100 80GB GPU.
The value of the threshold $s$ ranges from 0 to 1; a smaller $s$ results in more samples requiring front-door adjustment and consequently increases the cost of FLOPs. 
To facilitate a fair comparison of equivalent costs, we set the number of votes for CoT-SC at $m=1,5,8,10,40,80,100$. 
Our method achieves \textbf{superior} performance at equivalent costs.
As computational costs increase, the performance of CoT-SC gradually reaches its upper limit, whereas the performance of our proposed method continues to rise.
It indicates Causal Prompting has more potential to scale effectively with increased computational costs.

\paragraph{DQ2: How does adjusting the threshold $s$ affect performance?\\}

\begin{wrapfigure}{r}{8.5cm}
\vspace{-5mm}
 \includegraphics[width=0.48\textwidth]{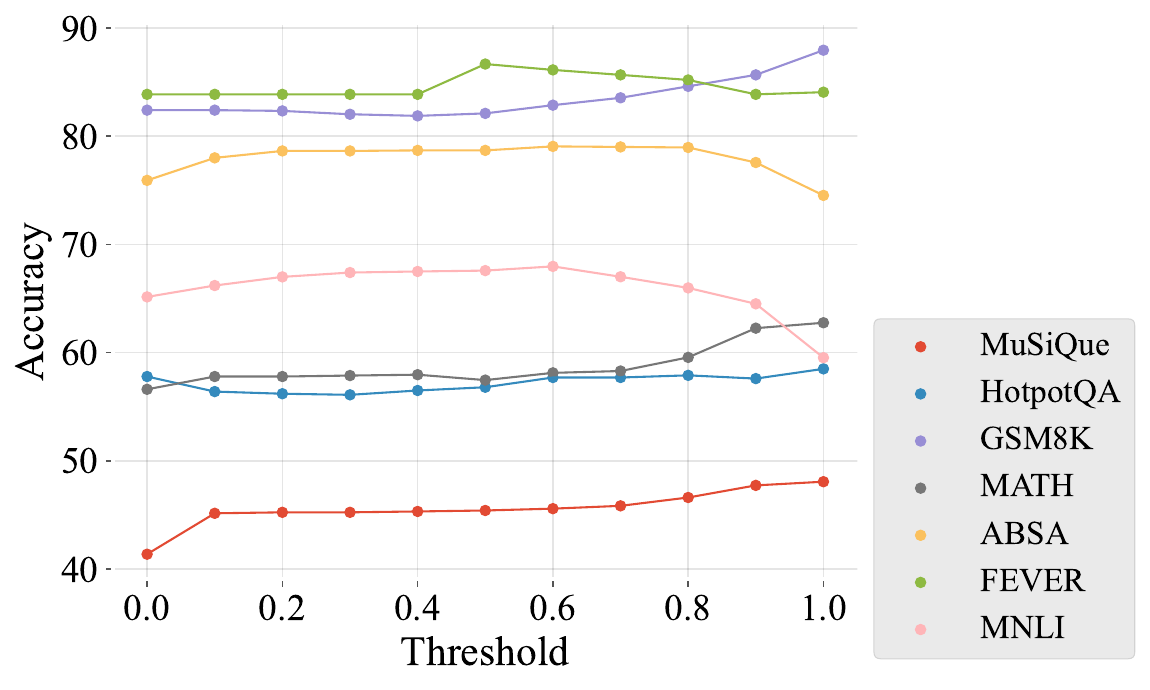}
 \caption{The impact of threshold $s$ on LLaMA3-8B.}
 \label{fig:threshold}
\vspace{-5mm}
\end{wrapfigure}
We also explored the effect of the threshold $s$ on different tasks on LLaMA3. 
As shown in Figure~\ref{fig:threshold}, the performance of mathematical reasoning and multi-hop question answering tasks keeps improving as the threshold $s$ increases. On the other hand, the performance on the NLU task first increases and then decreases, and when we apply the front-door adjustment to all samples ($\ie$, $s=1.0$), the performance drops significantly. This is because we introduce the wrong and correct chain-of-thoughts in the NWGM algorithm part, and the labels of the two chain-of-thoughts are usually opposite, which is easy to cause LLM to change the correct answer into the wrong one, especially in classification tasks such as NLU. 
The best threshold $s$ across different backbone LLMs and different datasets are shown in Table~\ref{tab:best-threshold}. Table 1 of the submitted manuscript reports the model's performance when the threshold $s$ takes the best value.

\paragraph{DQ3: What are the effects of employing contrastive learning methods?\\}

\begin{wrapfigure}{r}{8.5cm}
 \includegraphics[width=0.48\textwidth]{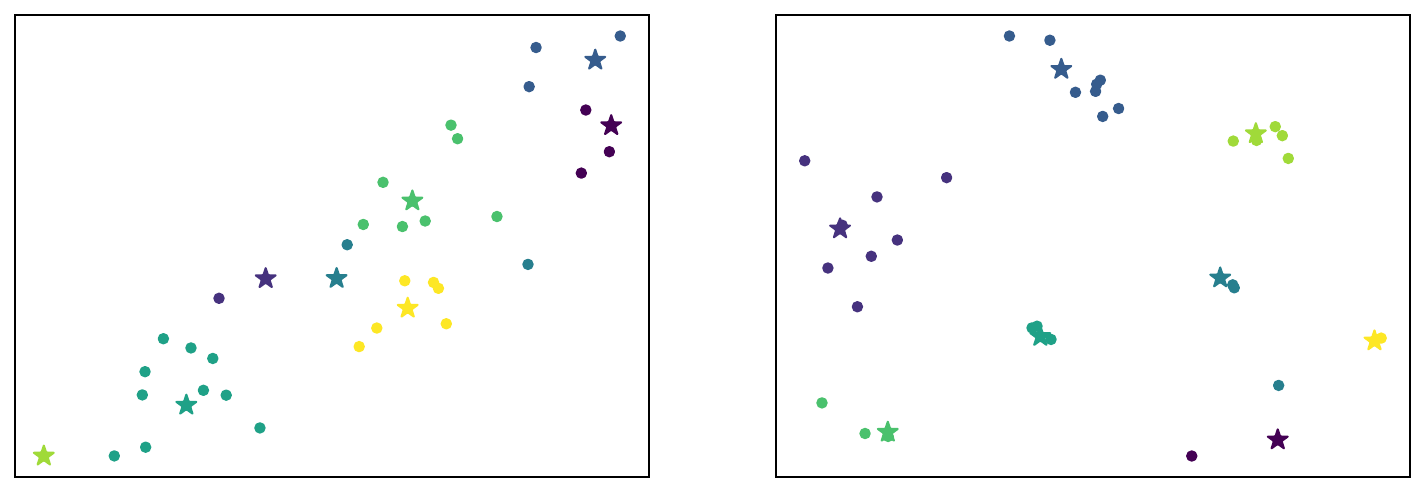}
 \caption{Visualization of the embeddings of chain-of-thought obtained by the original $\Encoder$ (left) and the $\Encoder$ trained through contrastive learning (right) on the GSM8K dataset.}
 \label{fig:cluster_visualization}
\end{wrapfigure}

To better explore the importance of contrastive learning, we present a visualization of the embeddings of the chain-of-thought obtained by the vanilla $\Encoder$ and the $\Encoder$ trained by contrastive learning with T-SNE~\citep{van2008visualizing}.
As shown in Figure~\ref{fig:cluster_visualization}, the vanilla $\Encoder$ fails to distinctly separate different categories, blending multiple chain-of-thoughts into indistinct representations.
In contrast, the $\Encoder$ that is trained by contrastive learning exhibits a clear delineation between different categories. 
This shows that contrastive learning can align the representation spaces of the encoder and LLMs, allowing the encoder to learn how to distinguish the semantics between different chain-of-thoughts generated by the LLMs.
In Appendix~\ref{sec:ablation}, we present quantitative analyses to further substantiate the effectiveness of contrastive learning.

\begin{table}[]
\small
\centering

\begin{tabular}{l|cc|cc|ccc}
\toprule
\textbf{Threshold $s$} & \textbf{GSM8K} & \textbf{MATH} & \textbf{HotpotQA} & \textbf{MuSiQue} & \textbf{ABSA} & \textbf{NLI} & \textbf{FV}  \\ \midrule
LLaMA2      & 1.0   & 0.2  & 0.5      & 1.0     & 0.4  & 0.4 & 0.6 \\
LLaMA3      & 1.0   & 1.0  & 1.0      & 1.0     & 0.6  & 0.6 & 0.5 \\
GPT-3.5     & 1.0   & 0.3  & 1.0      & 0.1     & 0.6  & 0.9 & 1.0 \\ \bottomrule
\end{tabular}%

\caption{The best threshold $s$ across different backbone LLMs, including LLaMA2, LLaMA3 and GPT-3.5, on seven datasets.}
\label{tab:best-threshold}
\end{table}
\paragraph{DQ4: Why do we not need any additional baselines?\\} 
See Appendix~\ref{sec:baseline}, where we list all the baselines used to compare with our proposed method. 
However, numerous other prompting strategies such as ToT~\citep{yao2024tree} or GoT~\citep{besta2024graph} have not been explored as baselines.
We do not conclude these strategies based on the following three primary considerations:
\textbf{First}, we focus on solving the more fundamental problem of prompting methods: the estimation of the causal effect between the input prompt and the response of the LLM. 
The efficacy of our causal prompting approach is substantiated by the experimental results presented.
\textbf{Second}, more complex prompting methods require the construction of intricate causal graphs.
In other words, building causal graphs for these advanced prompting methods is challenging due to their involvement with cycles and numerous mediating variables.
As an early work on combining causal inference with LLM, we start with the basic prompting method.
\textbf{Finally}, our approach is not specifically intended to improve the ability of LLMs to solve complex problems, but rather to hope that LLMs generate \textit{more causal, faithful and unbiased} answers. 
Therefore, we only experimented on the most commonly used tasks and did not compare them with other prompting methods designed for complex tasks.

\paragraph{DQ5: Why not consider the new bias from the CoTs generated by LLMs?\\}
In fact, we cannot guarantee that the CoTs generated by LLM do not introduce new biases.
In SCM, it is possible that $U$ points to $R$, and there is even the possibility of the $X-R$ link and the $R-A$ link being confounded by $U_1$ and $U_2$ respectively. 
However, it is very difficult to deal with such SCM, and following previous works~\citep{wu-etal-2024-decot,xu2015show,Tian_Cao_Zhang_Xing_2022,chen2023causal,zhang2024causal,yang2021deconfounded,zhu2023causal,wang-etal-2023-causal,xu-etal-2023-counterfactual,niu2021counterfactual}, we only make the assumption of a single confounder. 
We focus on the confounder between $X$ and $A$ and ignore the confounder of $R$ with other variables, aligning our causal graph with the front-door criterion. 
In practice, as shown in Figure 1 of the main body of the submitted manuscript, we observe that LLM can perform correct reasoning at the beginning, but it is often easily confused by bias in the last step of deriving the answer.
This shows the rationality of our assumed SCM. 

Additionally, our approach clusters multiple CoTs to capture diverse reasoning paths.
This diversity acts as a buffer, reducing the likelihood that any single biased pathway dominates the model's final decision. 
This is consistent with the principles of the front-door criterion, which relies on mediating variables to isolate the effect of the causal variable of interest. 

To further mitigate the confounder of $R$ with other variables, we implement the NWGM approximation by In-Context Learning.
Our NWGM algorithm enables LLM to improve the quality of the chain of thoughts by using ground truth demonstrations and maximally prevents the introduction of new bias when LLM generates the chain of thoughts.

In the future, we will try more complex SCM.

\section{Causal Prompting Algorithm}
\label{sec:algo}
\begin{table}[]
\small
\centering

\begin{tabular}{@{}c|c@{}}
\toprule
\textbf{Notation} & \textbf{Description} \\
\midrule
$\Encoder$ & The encoder-only model for generating text embedding \\
$\LLM$ & The large language model for text generation \\
$Sort$ & The function to sort the training set \\
$\mathcal{D}$ & The training set \\
$d$ & The demonstration examples in the prompt \\
$q^{test}$ & The test sample \\
$n$ & The number of demonstration examples in the prompt \\
$m$ & The number of CoTs generated by $\LLM$ \\
$K$ & The number of clusters for the K-means clustering algorithm \\
$T$ & The query times to $\LLM$ for each intervention prompt \\
\bottomrule
\end{tabular}
\caption{Notations used in our proposed method \textbf{Causal Prompting}.}
\label{tab:notations}
\end{table}
\subsection{Notations}
The notations used in the approach are shown in Table~\ref{tab:notations} to clarify their usage and significance throughout the algorithm.
\subsection{Algorithm Details}
As shown in Algorithm~\ref{alg:causal_prompting}, we describe the operation flow of Causal Prompting.

\begin{algorithm}[tb]
\caption{Causal Prompting}
\label{alg:causal_prompting}
\textbf{Input}: $\Encoder, \LLM, Sort, \mathcal{D}, d, q^{test}, n, m, K, T$
\begin{algorithmic}[1]
    \STATE $\mathcal{P} \gets [d_{1},...,d_{n},q^{test}]$
    \STATE $\{c_i|i=1,...,m\} \gets \LLM(\mathcal{P})$
    \STATE $\overline{c}_i \gets \Encoder(\text{\small [CLS]},c_i,\text{\small [SEP]})$ 
    \STATE $\{C_1,...,C_K\} \gets \Kmeans(\overline{c}_1,...,\overline{c}_m)$ 
    \STATE \textbf{for} $k=1$ \textbf{to} $K$:
    \STATE \quad $r_k \gets \Center(C_k)$
    \STATE \quad $P(r_k|do(X)) \gets \frac{|C_k|}{m}$ 
    \STATE \textbf{end for}
    \STATE \textbf{for} $k=1$ \textbf{to} $K$:
    \STATE \quad $\overline{r}_k \gets \Encoder(\text{\small [SEP]}, r_k,\text{\small [SEP]})$
    \STATE \quad $\overline{d}_j \gets \Encoder(\text{\small [CLS]},r_j^{wrong},\text{[\small SEP]})$
    \STATE \quad $\{d_j^{\uparrow}\}_{j=1}^N \gets Sort(\mathcal D, \overline{r}_k, \{\overline{d}_j\}_{j=1}^{N})$
    \STATE \quad $\mathcal{P}_{r_k}^{iter} \gets [d_l^{\uparrow},...,d_{1}^{\uparrow},q^{test}]$
    \STATE \quad $\{(r_{k,t}^{ip}, a_{k,t})| t=1,...,T\} \gets \LLM(\mathcal{P}_{r_k}^{iter},r_k)$
    \STATE \quad $P(A|do(r_k)) \gets \frac{\sum_{t=1}^T \mathbb{I}(A=a_{k,t})}{T}$
    \STATE \textbf{end for}
    \STATE $P(A|do(X)) \gets \sum_{k=1}^K \frac{|C_k|}{m} \cdot \frac{\sum_{t=1}^T \mathbb{I}(A=a_{k,t})}{T}$
    \RETURN $argmax_{A}(P(A|do(X)))$
\end{algorithmic}
\end{algorithm}

\section{Experimental Details}
\label{sec:exp}
\subsection{Baselines}
\label{sec:baseline}
\textbf{Standard ICL}~\citep{brown2020language}: Prompt LLMs with some demonstration examples containing only questions and their corresponding answers, without any additional explanatory context or reasoning.

\textbf{CoT}~\citep{wei2022chain}: Unlike Standard ICL, CoT method enhances the prompt with demonstration examples that include detailed chain-of-thoughts.
These chain-of-thoughts guide the LLMs through the steps required to reach an answer.

\textbf{CoT-SC}~\citep{wang2022self}: Extent the CoT methods by having the LLMs generate multiple different chain-of-thoughts for the same query and use majority voting to determine the final answer.

\subsection{Settings}
\label{sec:settings}
\paragraph{Demonstration Construction}
For the GSM8K and MATH datasets, we directly use the gold rationale in the dataset as the chain-of-thought for the demonstration. For multi-hop question answering and NLU tasks without gold rationale, we use a few manually constructed demonstrations to prompt the $\LLM$ to generate chain-of-thoughts and answers for all examples in the dataset. 
For the samples with wrong answers, we provide the $\LLM$ with the correct answers and then ask the $\LLM$ to generate the correct chain-of-thoughts. See Appendix~\ref{sec:prompt_demo_gen} for the prompt template used to generate the demonstration examples. Finally, we retain the wrong and correct chain-of-thoughts and use them to construct the training set.
Note that to better evaluate the debiasing effect of our method, we only use the original dataset to build demonstration examples without including the adversarial dataset, and evaluate on both original and adversarial datasets.

\paragraph{Demonstration Selection}
To fair comparison, the same demonstration samples are utilized across all few-shot prompting methods.
For each instance, the most relevant demonstration samples are selected based on the similarity of their embeddings with the question.
These selected demonstration samples are then concatenated into the prompt, as exemplified in the Appendix~\ref{sec:prompt_cot} for CoT prompting.
Specifically, for Math Reasoning and Multi-hop QA tasks, 4-shot and 2-shot settings are employed, respectively, corresponding to $n=4$ and $n=2$ in Equation 6 of the submitted manuscript. For NLU tasks, we maintain a balanced label space by including one demonstration sample for each category.
For ABSA and NLI, which are 3-way classification tasks, $n=3$ is adopted, while for FV, a 2-way task, $n=2$ is used.
After the application of NWGM-based causal intervention, as described in Section 3.2, and $l$ different demonstration samples are selected to be incorporated into the prompt as shown in the prompt templates of CoT Improvement based on NWGM approximation in Appendix~\ref{sec:prompt_causal_prompt}.
For the mathematical reasoning task, $l=2$ in Equation 16 of the submitted manuscript. 
For the multi-hop question answering task, $l=1$. 
For the ABSA and NLI tasks, $l=3$. For the FV task, we set $l=2$.

\subsection{Implementation Details}
\label{sec:implementation}
\paragraph{Details of $\LLM$}
We evaluate our prompting method on the open-source LLaMA-2-7B-Chat~\citep{touvron2023llama}\footnote{https://huggingface.co/meta-llama/Llama-2-7b-chat-hf} and LLaMA-3-8B-Instruct~\citep{llama3modelcard}\footnote{https://huggingface.co/meta-llama/Meta-Llama-3-8B-Instruct} using \textit{Transformer}\citep{wolf2019huggingface} library, as well as the closed-source \texttt{GPT-3.5-turbo-0125}~\citep{openaigpt35}\footnote{https://platform.openai.com/docs/models/gpt-3-5-turbo}.
The generation hyperparameters remain consistent across all prompting methods: temperature is set to $0.7$, and $top\_p$ is set to $0.9$. 
Following previous work~\citep{lyu-etal-2023-faithful}, we set the number of votes in COT-SC is $40$.
In our method, the number of chain-of-thoughts generated in the first part is $m=40$, and then these chain-of-thoughts are clustered into $K=8$. 
For each chain-of-thought representing the cluster center, we generated $T=10$ answers based on the prompt modified by intervention. 
Finally, the $K \cdot T=80$ answers were weighted voting to get the final answer.
\paragraph{Details of $\Encoder$}
We use BERT-base~\citep{devlin2018bert}\footnote{https://huggingface.co/google-bert/bert-base-uncased} as the $\Encoder$ in computing sentence similarity, clustering algorithm, and NWGM algorithm following~\citep{Tian_Cao_Zhang_Xing_2022,zhang2024causal}.
We independently fine-tune an $\Encoder$ for each $\LLM$, as well as for each specific task, employing a contrastive learning approach.
During training, we set the batch size is $128$. 
The learning rate is $1e-4$. The temperature $temp$ is set to $0.3$. The max length of $\Encoder$ is $512$. The total training epochs are $20$.
\subsection{Dataset Details}
\label{sec:dataset}

\paragraph{Math Reasoning}
For the GSM8K~\citep{cobbe2021training} dataset, we use its official dataset split\footnote{https://github.com/openai/grade-school-math}. The number of samples in the training set is 7473 and the number of samples in the test set is 1319.
For the MATH~\citep{hendrycksmath2021} dataset, we only use algebra type data due to limited computational resources. We use the official dataset split\footnote{https://github.com/hendrycks/math}, the training set of MATH-algebra includes 1744 samples, and the test set includes 1187 samples.

\paragraph{Multi-hop Question Answering}
For the HotpotQA~\citep{yang-etal-2018-hotpotqa} dataset, to reduce the experimental cost, we randomly selected 5000 samples from the official training set\footnote{http://curtis.ml.cmu.edu/datasets/hotpot/hotpot\_train\_v1.1.json} as our training set and 1000 samples from the official validation set\footnote{http://curtis.ml.cmu.edu/datasets/hotpot/hotpot\_dev\_distractor\_v1.json} as our test set.
For the MuSiQue~\citep{trivedi2021musique} dataset, we use the official dataset split of MuSiQue-Answerable version\footnote{https://github.com/StonyBrookNLP/musique}, and we experiment on the more challenging part of the dataset with $hop \ge 3$. After extracting the data with a $hop \ge 3$, the number of training sets is 5562 and the number of test sets is 1165.
For both datasets above, we extract the support documents to form the paragraph.

\paragraph{Natural Language Understanding}
For the Aspect-based Sentiment Analysis (ABSA) and Natural Language Inference (NLI) tasks, we use SemEval2014-Laptop~\citep{pontiki2016semeval} and MNLI-m~\citep{williams2017broad} as the original datasets (in-distribution, ID)  and the corresponding transformation data generated by TextFlint~\citep{wang2021textflint} as the adversarial datasets (out-of-distribution, OOD).
For the FV task, we use FEVER~\citep{thorne2018fever} as the ID dataset and its adversarial dataset Symmetric FEVER~\citep{schuster2019towards} as the OOD dataset.
To reduce the experimental cost, we randomly sample a certain number of samples from these NLU datasets for experiments.
See Table~\ref{tab:adv-desc} and Table~\ref{tab:data-statistic} for details of the adversarial dataset generation method and data statistics for the NLU dataset.

\begin{table}[]
\small
\centering

\begin{tabularx}{\textwidth}{ccX}
\toprule
\textbf{Task} &
  \textbf{Adversarial Category} &
 \textbf{Description} \\ \midrule
\multirow{4}{*}{ABSA} &
  ReverseTarget &
  Reverse the sentiment of the target aspect. \\
  & \multirow{2}{*}{ReverseNonTarget} &
  Reverse the sentiment of the non-target aspects with originally the same sentiment as target. \\
  & AddDiff &
  Add aspects with the opposite sentiment from the target aspect. \\ \hline
\multirow{5}{*}{NLI} &
  \multirow{2}{*}{AddSent} &
  Add some meaningless sentence to premise, which do not change the semantics. \\
 &
  \multirow{2}{*}{NumWord} &
  Find some num words in sentences and replace them with different num word. \\
 &
  SwapAnt &
  Find some keywords in sentences and replace them with their antonym.\\ \hline
  \multirow{3}{*}{FV} &
  \multirow{3}{*}{Symmetric} &
  For each claim-evidence pair, generating a synthetic pair that  holds the same relation (e.g. SUPPORTS or REFUTES) but  expressing a different, contrary, fact.  \\
\bottomrule 
\end{tabularx}
\caption{Multiple adversarial categories for ABSA, NLI, and FV tasks.}
\label{tab:adv-desc}
\end{table}

\begin{table}[]
\small
\centering

\begin{tabular}{@{}cccccc@{}}
\toprule
\multirow{2}{*}{\textbf{Tasks}}  & \multirow{2}{*}{\textbf{Datasets}} & \multirow{2}{*}{\textbf{Train}} & \multicolumn{2}{c}{\textbf{Test}} & \multirow{2}{*}{\textbf{Measure}}   \\
&          &       & \texttt{Ori}          & \texttt{Adv}         &              \\ \midrule
\multirow{2}{*}{Math Reasoning}               & GSM8K    & 7473  & 1319        & -          & Accuracy     \\
                                              & MATH     & 1744  & 1187        & -          & Accuracy     \\ \midrule
\multirow{2}{*}{Multi-hop Question Answering} & HotpotQA & 5000  & 1000        & -          & F1 \& EM \\
                                              & MuSiQue  & 5562  & 1165        & -          & F1 \& EM \\ \midrule
\multirow{3}{*}{Natural Language Understanding} & ABSA     & 2358  & 638         & 1239       & Accuracy     \\
                                              & NLI      & 5000  & 810         & 754        & Accuracy     \\
                                              & FV       & 5000  & 500         & 1000       & Accuracy     \\ \bottomrule
\end{tabular}

\caption{Details of all datasets used in the experiments.}
\label{tab:data-statistic}
\end{table}

\subsection{Evaluation}
\label{sec:eval}
Following previous works~\citep{ye2023comprehensive,lyu-etal-2023-faithful}, for Math Reasoning and NLU tasks, we adopt the label classification accuracy (Acc) as the evaluation metric, and for Multi-hop QA tasks, we adopt the Exact Match (EM) and F1~\citep{yang-etal-2018-hotpotqa} as the evaluation metric. Following previous work~\citep{lyu-etal-2023-faithful}, we extract the text span following the keyword "\textit{answer is}" as the answer.

\section{Related Works}
\label{sec:related_works}
\subsection{Prompting Strategies}
\label{sec:promptstra}
The performance of LLMs on downstream tasks is largely influenced by the employed prompting strategy~\citep{sahoo2024systematic}.
Current prompting strategies primarily improve the quantity and robustness through two approaches.
\textbf{On one hand}, incorporating a few labeled examples within the instruction can significantly enhance the performance of LLM, occasionally even better than fine-tuned models~\citep{brown2020language,chung2022scaling,dong2022survey}.
This method is referred to as In-Context Learning (ICL).
Several works involve the modification of instruction examples to address the issue of LLMs being particularly sensitive to the designing of demonstration examples, including example selection~\citep{liu2021makes}, example format~\cite{dong2022survey}, example label~\citep{min2022rethinking,yoo2022ground}, and example order~\citep{lu2021fantastically}. 
Following this way, recent works have suggested that including chain-of-thoughts (CoTs) in the context of these examples can further augment the quality of LLM's responses~\citep{nye2021show,lampinen2022can,wei2022chain}.
\textbf{On the other hand}, to tackle the instability in the outputs of LLMs, attributable to the \textit{``multinomial sampling''}~\citep{holtzman2019curious}, some works involve enhancement by employing multiple sampling and voting mechanisms to determine the final answer~\citep{chen2021evaluating,wang2022self,li2022advance}, which is referred to as self-consistency (SC). Moreover, for more complex tasks, many other prompting methods based on multiple sampling have been proposed, such as React~\citep{yao2022react}, ToT~\citep{yao2024tree}, and GoT~\citep{besta2024graph}.
 
In this work, we primarily combine these two directions by modeling front-door adjustments to address bias issues in LLM prompting. 
CoT is utilized as a mediator variable.
In such scenarios, we further employ NWGM approximation to select the demonstration, which can represent the expectation of the entire dataset and help the model generate an unbiased answer.

\subsection{Debiasing with Causal Inference}
Causal inference uses scientific methods to identify causal relationships between variables~\citep{pearl2016causal}. 
Because of its rigorous theoretical guarantees and mature causal modeling tools~\citep{pearl2019seven}, causal inference has advantages in debiasing work. 
Recently, causal inference has been widely used in natural language processing~\citep{feder2022causal, chen-etal-2023-causal} and computer vision~\citep{yang2021deconfounded,wu2024unlearning,wu2024video,wu2024number,qin2024pearlinputagnosticpromptenhancement}. 

Some works apply counterfactual reference to remove the bias of the model~\citep{xu-etal-2023-counterfactual,guo-etal-2023-counterfactual,niu2021counterfactual,xu2024take}. 
For instance, \citet{niu2021counterfactual} debias the visual question answering task by subtracting the predictions of the language-only model from the predictions of the vision-language model to reduce linguistic biases in the integrated system.
Other work uses causal interventions for debiasing, including backdoor adjustment~\citep{Tian_Cao_Zhang_Xing_2022,zhu2023causal,wang-etal-2023-causal,wu2024diner} and front-door adjustment~\citep{yang2021causal,yang2021deconfounded,zhang2024causal}. 
In the era of LLMs, several studies have integrated LLMs with causal inference techniques~\citep{jin2023cladder,lyu2024causal,jin2023can,stolfo-etal-2023-causal}. 
However, some of these studies do not fully adhere to the structural causal model, relying excessively on heuristic methods~\citep{lu2022neuro,wang-etal-2023-causal,zhang2023mitigating,tang2023towards}; 
others employ overly simplistic causal diagrams, which are inadequate for complex tasks~\citep{abdali2023extracting,lyu2024causal}.
Overall, counterfactual inference necessitates the acquisition of logits from LLM outputs, whereas back-door adjustment demands modeling of specific values of confounding variables.
In contrast, front-door adjustment allows causal intervention without access to the values of confounding variables or logits of LLM outputs. 
This makes front-door adjustment particularly apt for application in LLM scenarios.

Therefore, we propose to debias the prompting methods by causal intervention based on front-door adjustment. 
The work most closely related to ours is DeCOT~\citep{wu-etal-2024-decot};
DeCoT~\citep{wu-etal-2024-decot} debiases the chain-of-thoughts of LLMs by incoroporating counterfactual knowledge and front-door adjustment. 
Both our work and DeCoT use the front-door adjustment on LLMs.
Howerver, DeCoT requires the introduction of instrumental variables to model counterfactual knowledge, limiting its applicability to knowledge-intensive tasks.
In contrast, our approach is versatile and can be applied to a wide range of tasks.
Our approach, on one hand, adapts the traditional front-door adjustment to make it suitable for the task of LLM prompting. On the other hand, it adheres more closely to the established principles of the field.


\section{More Experimental Results}
\subsection{Robustness Study}
\label{sec:more_robustness}
\begin{table}[]
\centering
\small

\begin{tabular}{@{}ccccccc@{}}
\toprule
\multicolumn{1}{c|}{} &
  \multicolumn{2}{c|}{ABSA} &
  \multicolumn{2}{c|}{NLI} &
  \multicolumn{2}{c}{FV} \\ \midrule
\multicolumn{1}{c|}{Methods} &
 \texttt{Ori} &
  \multicolumn{1}{c|}{\texttt{Adv}} &
  \texttt{Ori} &
  \multicolumn{1}{c|}{\texttt{Adv}} &
  \texttt{Ori} &
  \texttt{Adv} \\ \midrule
\rowcolor{mygray} \multicolumn{7}{c}{ LLaMA2} \\ \midrule
\multicolumn{1}{c|}{Standard ICL} &
  71.63 &
  \multicolumn{1}{c|}{34.71} &
  36.42 &
  \multicolumn{1}{c|}{19.36} &
  72.00 &
  49.30 \\
\multicolumn{1}{c|}{CoT} &
  66.14 &
  \multicolumn{1}{c|}{40.36} &
  29.51 &
  \multicolumn{1}{c|}{25.46} &
  76.60 &
  66.80 \\
\multicolumn{1}{c|}{CoT-SC} &
  \textbf{75.24} &
  \multicolumn{1}{c|}{42.62} &
  36.91 &
  \multicolumn{1}{c|}{29.97} &
  77.80 &
  69.40 \\
\multicolumn{1}{c|}{Causal Prompting} &
  73.04 &
  \multicolumn{1}{c|}{\textbf{64.73}} &
  \textbf{55.06} &
  \multicolumn{1}{c|}{\textbf{46.29}} &
  \textbf{89.00} &
  \textbf{77.10} \\ \midrule
\rowcolor{mygray} \multicolumn{7}{c}{GPT-3.5} \\ \midrule
\multicolumn{1}{c|}{Standard ICL} &
  75.24 &
  \multicolumn{1}{c|}{66.18} &
  74.32 &
  \multicolumn{1}{c|}{31.17} &
  88.20 &
  68.90 \\
\multicolumn{1}{c|}{CoT} &
  70.69 &
  \multicolumn{1}{c|}{63.2} &
  74.07 &
  \multicolumn{1}{c|}{52.25} &
  89.60 &
  76.20 \\
\multicolumn{1}{c|}{CoT-SC} &
  79.94 &
  \multicolumn{1}{c|}{71.83} &
  76.91 &
  \multicolumn{1}{c|}{56.10} &
  91.00 &
  78.60 \\
\multicolumn{1}{c|}{Causal Prompting} &
  \textbf{82.29} &
  \multicolumn{1}{c|}{\textbf{79.02}} &
  \textbf{80.74} &
  \multicolumn{1}{c|}{\textbf{62.47}} &
  \textbf{94.60} &
  \textbf{82.50} \\ \bottomrule
\end{tabular}%

\caption{Results of the robustness study on LLaMA2 and GPT-3.5. \texttt{Ori} denotes the original dataset (in-distribution) and \texttt{Adv} denotes the adversarial dataset (out-of-distribution). The best results are in \textbf{bold}.}
\label{tab:results-robustness-all}
\end{table}
We also provide robustness studies on LLaMA2 and GPT3.5 in Table~\ref{tab:results-robustness-all}.
The findings are consistent with experiments conducted on LLaMA3 in Section 4.4, demonstrating that our method possesses a distinct advantage on adversarial robustness datasets.

\subsection{Ablation Study}
\label{sec:ablation}
\begin{table}[]
\small
\centering

\begin{tabular}{@{}l|cc|cccc@{}}
\toprule
                         & GSM8K & MATH  & \multicolumn{2}{c}{HotpotQA} & \multicolumn{2}{c}{MuSiQue} \\ \midrule
Methods                  & Acc   & Acc   & EM           & F1            & EM           & F1           \\ \midrule
Causal Prompting & \textbf{87.95} & \textbf{62.76} & \textbf{58.5} & \textbf{78.18} & \textbf{48.07} & \textbf{64.23} \\
\quad NWGM-\texttt{Reverse}             & 87.72 & 62.26 & 57.7         & 77.97         & 47.81        & 63.48        \\
\quad NWGM-\texttt{Random}              & 86.13 & 52.99 & 56.9         & 77.23         & 47.38        & 61.60         \\
\quad w/o Contrastive Learning & 86.58 & 59.56 & 57.9         & 77.88         & 47.47        & 62.45        \\ 
\quad w/o K-means     & 84.46 & 56.7  & 56.4 & 77.2  & 47.12 & 61.63 \\ 
\quad w/o Weighting   & 81.35 & 45.07 & 55.1 & 75.98 & 41.29 & 56.83 \\
\bottomrule
\end{tabular}

\caption{The results of ablation study on LLaMA3. The best results are in \textbf{bold}.}
\label{tab:results-ablation}
\end{table}
In the ablation study presented in Table~\ref{tab:results-ablation}, we perform a detailed ablation analysis to evaluate three pivotal aspects: (1) the effectiveness of the NWGM approximation, (2) the impact of incorporating contrastive learning, (3) the impact of K-means clustering and weighting mechanism.

This analysis spans four datasets, including GSM8K, MATH, HotpotQA, and MuSiQue, utilizing the LLaMA3-8B model. 
The comparison results between Causal Prompting and the baselines (Table 1 of the submitted manuscript) and the results of robustness studies (Tables 2 of the submitted manuscript) demonstrate that the Causal Prompting method consistently exhibits superior performance across GPT-3.5, LLaMA3, and LLaMA2, showing high performance across various test metrics.
This indicates that the Causal Prompting method possesses excellent generalizability and stability. 
Since GPT-3.5 and LLaMA3 perform consistently, we only perform ablation experiments on LLaMA3 in order to reduce the economic cost.

\textbf{Effectiveness of the NWGM}
The NWGM approximation is employed in the Estimation of $P(A|do(r))$ to perform the back-door adjustment, where we select the ICL demonstration that is most similar to the query and put the most similar demonstration samples closer to the test samples, as detailed in Section 3.2.
We evaluate the impact of the NWGM approximation by comparing the standard setup against two variants of NWGM-\texttt{Reverse} and NWGM-\texttt{Random}, respectively. NWGM-\texttt{Reverse} means that we reverse the order of standard ICL demonstrations, that is, $\mathcal{P}_{r_k}^{iter} = [d_1^{\uparrow},...,d_{n}^{\uparrow},q^{test}]$ with the same order of KATE~\citep{liu2021makes} method.
NWGM-\texttt{Random} denotes that ICL demonstrations are selected at random from the training set $\mathcal D$.
The performance decline observed in these two variants validates the effectiveness of our approach in selecting the most relevant samples, thereby enabling a more accurate estimation of $P(A|do(r))$.

\textbf{Impact of contrastive learning}
The second aspect of our ablation study is to assess the role of contrastive learning.
By removing contrastive learning, we can observe a decline in performance metrics compared to Causal Prompting.
For example, there is a decrease of 1.37\% in accuracy on GSM8K and 3.20\% on MATH.
It indicates that contrastive learning facilitates the alignment of feature representations between the $Encoder$ and LLMs.

\textbf{Impact of K-means clustering and weighting}
The third aspect of our ablation study is to explore the contribution of K-means clustering, and the weighting mechanism.
w/o K-means means that K chain-of-thoughts are randomly selected in the first stage of front-door adjustment, and $P(r|do(X))=1/K$ is set. w/o Weighting means that instead of using $P(r|do(X))$ and $P(A|do(r))$ to weighted sum the final $K*T$ answers, a majority vote is taken on these answers. 
Regarding the influence of K-means clustering, removing the K-means clustering will destroy the estimate of $P(r|do(X))$ in the front-door adjustment, leading to a decrease in model performance. 
Removing weighting destroys both the estimation of $P(r|do(X))$ and $P(A|do(r))$ in the front-door adjustment, leading to a significant decrease in model performance.

\subsection{Hyperparameter Study}
\label{sec:hyper_study}

\begin{table}[]
\centering

\begin{tabular}{c|cc|cccc}
\hline
            & GSM8K          & MATH           & \multicolumn{2}{c}{HotpotQA}   & \multicolumn{2}{c}{MuSiQue}     \\ \hline
Cluster Num & Acc            & Acc            & EM            & F1             & EM             & F1             \\ \hline
1           & 83.4           & 44.14          & 55.6          & 76.18          & 32.19          & 50.11          \\
4           & 86.35          & 47.77          & 56.4          & 77.06          & 43.18          & 59.43          \\
8           & 87.95          & \textbf{62.76} & 58.50         & 78.18          & 48.07          & \textbf{64.23} \\
12          & \textbf{88.55} & 62.26          & \textbf{58.7} & \textbf{78.49} & 48.07          & 63.02          \\
16          & 88.25          & 59.56          & 58.1          & 78.15          & 48.67          & 62.7           \\
20          & 88.32          & 58.3           & 58.50         & 78.45          & \textbf{49.79} & 63.81          \\ \hline
\end{tabular}
\caption{The results of hyperparameters for the number of clusters. The chain of thoughts generated in the first stage is 40. The best results are in bold.}
\label{tab:results-hyper-cluster-num}
\end{table}

\begin{table}[]
\centering

\begin{tabular}{c|cc|cccc}
\hline
\textbf{}   & GSM8K          & MATH           & \multicolumn{2}{c}{HotpotQA}   & \multicolumn{2}{c}{MuSiQue}   \\ \hline
CoT Num & Acc            & Acc            & EM            & F1             & EM            & F1            \\ \hline
8           & 82.11          & 54.84          & 56.1          & 76.83          & 42.4          & 58.17         \\
20          & 84.61          & 56.19          & 56.8          & 77.53          & 46.35         & 60.49         \\
40          & 87.95          & \textbf{62.76} & 58.50         & 78.18          & 48.07         & 64.23         \\
60          & 87.87          & 62.34          & 58.1          & 78.44          & 48.33         & 62.78         \\
80          & 88.4           & 62.68          & \textbf{58.6} & 78.67          & 48.07         & 62.3          \\
100         & \textbf{88.93} & 61.75          & \textbf{58.6} & \textbf{78.86} & \textbf{49.1} & \textbf{64.8} \\ \hline
\end{tabular}
\caption{The results of hyperparameters for the number of chain of thoughts. The cluster num in the second stage is 8. The best results are in bold.}
\label{tab:results-hyper-cot-num}
\end{table}
We conduct additional hyperparameter experiments to explore the impact of the number of clusters $K$ and the number of CoTs $m$ on the performance.

\textbf{Impcat of the number of clusters $K$}
As shown in Table~\ref{tab:results-hyper-cluster-num}, we conducted further experiments with varying cluster numbers and included an analysis of how the choice of cluster number influences the results.
It can be observed that when the number of clusters increases from 1 to 8, the performance improvement is obvious, and from 8 to 20, the performance remains stable or even decreases. Since the number of chain-of-thoughts $m$ is limited, $P(r|do(X))$ cannot be accurately estimated when the number of clusters is too small or too large.
Therefore, considering both performance and cost, we choose the number of clusters $K=8$.

\textbf{Impact of the number of CoTs $m$}
As shown in Table~\ref{tab:results-hyper-cot-num}, we show the performance with different numbers of chain-of-thoughts. It can be found that when the number of CoTs increases from 8 to 40, the performance improvement is significant, and when the number of CoTs increases from 40 to 100, the performance improvement is slight. It demonstrates that the number of chain of thought $m=40$ is already sufficient to estimate $P(r|do(X))$. 
Therefore, considering the cost issue, we choose the number of CoTs $m=40$.

\section{Limitations}
\label{sec:limit}
Although our results already outperform baselines overall, our work still suffers from the following limitations.
\begin{itemize}
    \item We evaluate the effectiveness of our approach on three tasks: Math Reasoning, Multi-hop Question Answering, and Natural Language Understanding. We need to test the effectiveness of Causal Prompting on other more complex tasks.
    \item As mentioned in Appendix~\ref{sec:dis}, although our method outperforms baselines at the same cost, how to reduce the cost of prompt remains an important issue.
    \item We only evaluated the effectiveness of Causal Prompting on three Large Language Models, LLaMA2-7B, LLaMA3-8B and GPT-3.5, and we need to evaluate our method on more Large Language Models of different kinds and scales.
\end{itemize}

\section{Case Study}
\label{sec:case_study}
In this section, we provide two running samples from GSM8K and HotpotQA with intermediate output for each module.
\subsection{Case on GSM8K}
\phantomsection
\label{case:case_GSM8K}
\begin{tcolorbox}[enhanced jigsaw,breakable,pad at break*=1mm,colback=white!95!gray,colframe=blue!50!black,rounded corners,label={template-second-select-roles}, title={Intermediate output details of a GSM8K example.}]
\textbf{Question}
\begin{lstlisting}[breaklines=true, xleftmargin=0pt, breakindent=0pt, columns=fullflexible]
A merchant wants to make a choice of purchase between 2 purchase plans: jewelry worth $5,000 or electronic gadgets worth $8,000. His financial advisor speculates that the jewelry market will go up 2.5% while the electronic gadgets market will rise 1.2% within the same month. If the merchant is looking to maximize profit at the end of this month by making a choice, how much profit would this be? 
(*@\textcolor{c2}{(The correct answer is 125)}@*)
\end{lstlisting}
\textbf{Outputs from CoT-SC}
\begin{lstlisting}[breaklines=true, xleftmargin=0pt, breakindent=0pt, columns=fullflexible]
The number of chain-of-thoughts m=40.
Answer voting results:
(*@$P(A=96|X) = 19/40 = 0.475$@*)
(*@$P(A=125|X) = 18/40 = 0.45$@*)
...
(Omit answers with fewer votes)
Therefore, the final answer obtained according to the COT-SC method is (*@\textcolor{c3}{96}@*).

An example of an error is shown:

Let's calculate the future value of each option:
Option 1: Jewelry
Current value = $5,000
Expected increase = 2.5% of $5,000 = $125
Future value = $5,000 + $125 = $5,125
Option 2: Electronic Gadgets
Current value = $8,000
Expected increase = 1.2% of $8,000 = $96
Future value = $8,000 + $96 = $8,096
Since the merchant wants to maximize profit, they should choose the option with the higher future value. In this case, it's Option 2: Electronic Gadgets with a future value of $8,096.
The profit would be the difference between the future value and the current value: $8,096 - $8,000 = $96.
Therefore, the answer is $(*@\textcolor{c3}{96}@*).
\end{lstlisting}
\textbf{Cluster results}
\begin{lstlisting}[breaklines=true, xleftmargin=0pt,breakindent=0pt, columns=fullflexible]
The number of chain-of-thoughts selected from clustering results k=8.
--------------------
CoT-1: (*@$P(r_1|X) = 0.375$@*)
--------------------
Let's calculate the profit for each option:
Option 1: Jewelry
* Current value: $5,000
* Expected increase: 2.5% of $5,000 = $125
* Future value: $5,000 + $125 = $5,125
* Profit: $5,125 - $5,000 = $125
Option 2: Electronic Gadgets
* Current value: $8,000
* Expected increase: 1.2% of $8,000 = $96
* Future value: $8,000 + $96 = $8,096
* Profit: $8,096 - $8,000 = $96
Comparing the profits, the merchant would make a higher profit by choosing the jewelry option, which is $125. Therefore, the answer is $125.

--------------------
CoT-2: (*@$P(r_2|X) = 0.025$@*)
--------------------
Let's analyze the situation step by step:
1. The merchant has to choose between jewelry worth $5,000 and electronic gadgets worth $8,000.
2. The financial advisor predicts that the jewelry market will go up 2.5% within the same month. This means that the value of the jewelry will increase by 2.5% of $5,000, which is:
$5,000 x 0.025 = $125
So, the value of the jewelry will become:
$5,000 + $125 = $5,125
3. The financial advisor predicts that the electronic gadgets market will rise 1.2% within the same month. This means that the value of the electronic gadgets will increase by 1.2% of $8,000, which is:
$8,000 x 0.012 = $96
So, the value of the electronic gadgets will become:
$8,000 + $96 = $8,096
4. Now, let's compare the profit from each option:
Option 1: Jewelry ($5,125 - $5,000) = $125
Option 2: Electronic Gadgets ($8,096 - $8,000) = $96
Since $125 is greater than $96, the merchant would make a maximum profit by choosing the jewelry option, which is $125.

--------------------
CoT-3: (*@$P(r_3|X) = 0.075$@*)
--------------------
Let's analyze the situation:
Option 1: Jewelry worth $5,000
* The jewelry market will go up 2.5%, so the value will increase by 2.5% of $5,000 = $125
* The merchant will sell the jewelry for $5,000 + $125 = $5,125
* The profit = Selling price - Cost price = $5,125 - $5,000 = $125
Option 2: Electronic gadgets worth $8,000
* The electronic gadgets market will go up 1.2%, so the value will increase by 1.2% of $8,000 = $96
* The merchant will sell the electronic gadgets for $8,000 + $96 = $8,096
* The profit = Selling price - Cost price = $8,096 - $8,000 = $96
Comparing the profits, we can see that the merchant would make a profit of $125 if they choose the jewelry option, and a profit of $96 if they choose the electronic gadgets option. Since $125 is greater than $96, the merchant would maximize their profit by choosing the jewelry option.

--------------------
CoT-4: (*@$P(r_4|X) = 0.025$@*)
--------------------
Let's analyze the situation step by step:
1. If the merchant chooses to buy jewelry worth $5,000, its value will increase by 2.5% in the next month. To find the increase in value, multiply the current value by the percentage increase:
$5,000 x 2.5% = $5,000 x 0.025 = $125
The total value of the jewelry after the increase will be:
$5,000 + $125 = $5,125
2. If the merchant chooses to buy electronic gadgets worth $8,000, its value will increase by 1.2% in the next month. To find the increase in value, multiply the current value by the percentage increase:
$8,000 x 1.2% = $8,000 x 0.012 = $96
The total value of the electronic gadgets after the increase will be:
$8,000 + $96 = $8,096
3. To maximize profit, the merchant should choose the option that results in the highest total value. In this case, the merchant should choose to buy electronic gadgets worth $8,096, which is $976 more than the value of the jewelry.
The profit will be the difference between the increased value of the electronic gadgets and their original price:
$8,096 - $8,000 = $96
So, the maximum profit the merchant can make is $96.

--------------------
CoT-5: (*@$P(r_5|X) = 0.15$@*)
--------------------
Let's calculate the value of the jewelry after the market goes up 2.5%:
Initial value: $5,000
Percentage increase: 2.5%
Value after increase: $5,000 + (2.5% of $5,000) = $5,000 + $125 = $5,125
Now, let's calculate the value of the electronic gadgets after the market goes up 1.2%:
Initial value: $8,000
Percentage increase: 1.2%
Value after increase: $8,000 + (1.2% of $8,000) = $8,000 + $96 = $8,096
Since the merchant wants to maximize profit, he should choose the option that gives him the highest value after the market increase. In this case, the electronic gadgets option is worth $8,096, which is higher than the jewelry option worth $5,125.
The profit would be the difference between the value of the chosen option and its initial value:
Profit = $8,096 - $8,000 = $96
Therefore, the answer is $96.

--------------------
CoT-6: (*@$P(r_6|X) = 0.025$@*)
--------------------
Let's analyze the situation step by step:
1. The merchant has two options: purchasing jewelry worth $5,000 or electronic gadgets worth $8,000.
2. The financial advisor predicts that the jewelry market will rise by 2.5%, which means the value of the jewelry will increase by 2.5% of $5,000 = $125.
3. The value of the jewelry will then become $5,000 + $125 = $5,125.
4. The financial advisor also predicts that the electronic gadgets market will rise by 1.2%, which means the value of the electronic gadgets will increase by 1.2% of $8,000 = $96.
5. The value of the electronic gadgets will then become $8,000 + $96 = $8,096.
6. The merchant wants to maximize profit, which means they should choose the option that results in the highest profit.
Comparing the two options:
* Jewelry: $5,125 (new value) - $5,000 (original value) = $125 profit
* Electronic gadgets: $8,096 (new value) - $8,000 (original value) = $96 profit
The merchant should choose the jewelry option, as it results in a higher profit of $125 compared to $96 for the electronic gadgets. Therefore, the answer is $125.

--------------------
CoT-7: (*@$P(r_7|X) = 0.15$@*)
--------------------
Let's calculate the profit for each option:
Option 1: Jewelry worth $5,000
The jewelry market will rise 2.5%, so the value of the jewelry will increase to:
$5,000 x (1 + 0.025) = $5,000 x 1.025 = $5,125
Profit = Selling price - Purchase price = $5,125 - $5,000 = $125
Option 2: Electronic gadgets worth $8,000
The electronic gadgets market will rise 1.2%, so the value of the gadgets will increase to:
$8,000 x (1 + 0.012) = $8,000 x 1.012 = $8,096
Profit = Selling price - Purchase price = $8,096 - $8,000 = $96
Since the merchant is looking to maximize profit, they should choose Option 1, which yields a profit of $125.

--------------------
CoT-8: (*@$P(r_8|X) = 0.175$@*)
--------------------
Let's analyze the situation step by step:
1. Jewelry worth $5,000 with a 2.5% increase:
New value = $5,000 + (2.5% of $5,000) = $5,000 + $125 = $5,125
Profit = New value - Original value = $5,125 - $5,000 = $125
2. Electronic gadgets worth $8,000 with a 1.2% increase:
New value = $8,000 + (1.2% of $8,000) = $8,000 + $96 = $8,096
Profit = New value - Original value = $8,096 - $8,000 = $96
Comparing the profits, the merchant would make a profit of $125 if they choose the jewelry plan and a profit of $96 if they choose the electronic gadgets plan. Since the merchant wants to maximize profit, they should choose the jewelry plan, which would yield a profit of $125.

\end{lstlisting}
\textbf{Intervention results}
\begin{lstlisting}[breaklines=true, xleftmargin=0pt,breakindent=0pt, columns=fullflexible]
For the above eight chain-of-thoughts, we use the NWGM algorithm to perform causal intervention operations on them respectively. We can then compute the causal effect between the chain-of-thought and the answer.

(*@$P(A=125|do(r_1)) = 9/10 = 0.9$@*)
(*@$P(A=96|do(r_1)) = 0/10 = 0.0$@*)

(*@$P(A=125|do(r_2)) = 10/10 = 1.0$@*)
(*@$P(A=96|do(r_2)) = 0/10 = 0.0$@*)

(*@$P(A=125|do(r_3)) = 10/10 = 1.0$@*)
(*@$P(A=96|do(r_3)) = 0/10 = 0.0$@*)

(*@$P(A=125|do(r_4)) = 0/10 = 0.0$@*)
(*@$P(A=96|do(r_4)) = 8/10 = 0.8$@*)

(*@$P(A=125|do(r_5)) = 0/10 = 0.0$@*)
(*@$P(A=96|do(r_5)) = 10/10 = 1.0$@*)

(*@$P(A=125|do(r_6)) = 9/10 = 0.9$@*)
(*@$P(A=96|do(r_6)) = 1/10 = 0.1$@*)

(*@$P(A=125|do(r_7)) = 8/10 = 0.8$@*)
(*@$P(A=96|do(r_7)) = 0/10 = 0.0$@*)

(*@$P(A=125|do(r_8)) = 10/10 = 1.0$@*)
(*@$P(A=96|do(r_8)) = 0/10 = 0.0$@*)

\end{lstlisting}

\textbf{Final results}
\begin{lstlisting}[breaklines=true, xleftmargin=0pt,breakindent=0pt, columns=fullflexible]
The final answer is obtained by performing a weighted voting as follows:
(*@$P(A=125|do(X))$@*) = 0.375 * 0.9 + 0.025 * 1.0 + 0.075 * 1.0 + 0.025 * 0.0 + 0.15 * 0.0 + 0.025 * 0.9 + 0.15 * 0.8 + 0.175 * 1.0 = 0.755

(*@$P(A=96|do(X))$@*) = 0.375 * 0.0 + 0.025 * 0.0 + 0.075 * 0.0 + 0.025 * 0.8 + 0.15 * 1.0 + 0.025 * 0.1 + 0.15 * 0.0 + 0.175 * 0.0 = 0.1725
Finally, we chose the answer with the largest weight as the final answer.
Therefore, the final answer obtained according to the Causal Prompting method is (*@\textcolor{c2}{125}@*).
\end{lstlisting}
\end{tcolorbox}

\subsection{Case on HotpotQA}
\phantomsection
\label{case:case_HotpotQA}
\begin{tcolorbox}[enhanced jigsaw,breakable,pad at break*=1mm,colback=white!95!gray,colframe=blue!50!black,rounded corners,label={template-second-select-roles}, title={Intermediate output details of a HotpotQA example.}]
\textbf{Context}
\begin{lstlisting}[breaklines=true, xleftmargin=0pt, breakindent=0pt, columns=fullflexible]
The Walton family is an American family whose collective fortune makes them among the richest families in the world.  The majority of their wealth derives from the heritage of Bud and Sam Walton, who were the co-founders of the world's largest retailer, Walmart.  The three most prominent living members (Jim, Rob and Alice) have consistently been in the top ten of the "Forbes" 400 since 2001, as were John ( 2005) and Helen (d. 2007) prior to their deaths.  Christy Walton took her husband John's place after his death. Helen Robson Kemper Walton (December 3, 1919 - April 19, 2007) was the wife of Wal-Mart and Sam's Club founder Sam Walton.  At one point in her life, she was the richest American and the eleventh-richest woman in the world.
\end{lstlisting}
\textbf{Question}
\begin{lstlisting}[breaklines=true, xleftmargin=0pt, breakindent=0pt, columns=fullflexible]
Who was the husband of the prominent Walton family member who died after John died in 2005? 
(*@\textcolor{c2}{(The correct answer is Sam Walton)}@*)
\end{lstlisting}
\textbf{Outputs from CoT-SC}
\begin{lstlisting}[breaklines=true, xleftmargin=0pt, breakindent=0pt, columns=fullflexible]
The number of chain-of-thoughts m=40.
Answer voting results:
(*@$P(A=John|X) = 13/40 = 0.325$@*)
(*@$P(A=Sam Walton|X) = 12/40 = 0.3$@*)
(*@$P(A=John Walton|X) = 9/40 = 0.225$@*)
...
(Omit answers with fewer votes)
Therefore, the final answer obtained according to the COT-SC method is (*@\textcolor{c3}{John}@*).

An example of an error is shown:

Step 1: Identify the prominent Walton family member who died after John died in 2005. Helen Robson Kemper Walton died in 2007.
Step 2: Determine who Helen was married to. Helen was the wife of Sam Walton.
Step 3: Identify who took Helen's place after her death. Christy Walton took her husband John's place after his death.
Step 4: Determine who John is. John is the husband of Christy Walton.
Step 5: Combine the information to answer the question. The husband of the prominent Walton family member who died after John died in 2005 is John.
Therefore, the answer is: (*@\textcolor{c3}{John}@*).
\end{lstlisting}
\textbf{Cluster results}
\begin{lstlisting}[breaklines=true, xleftmargin=0pt,breakindent=0pt, columns=fullflexible]
The number of chain-of-thoughts selected from clustering results k=8.
--------------------
CoT-1: (*@$P(r_1|X) = 0.325$@*)
--------------------
Step 1: Identify the prominent Walton family member who died after John died in 2005. According to the text, Helen Robson Kemper Walton died in 2007.
Step 2: Determine who Helen's husband was. According to the text, Helen was the wife of Sam Walton.
Step 3: Determine who John's wife was. According to the text, John's wife was Christy Walton.
Step 4: Combine the information to answer the question. Since Helen was Sam's wife and Christy is John's wife, it means that John is not Helen's husband. Therefore, the answer is: Sam Walton.

--------------------
CoT-2: (*@$P(r_2|X) = 0.25$@*)
--------------------
Step 1: Identify the prominent Walton family member who died after John died in 2005. The prominent Walton family member who died after John died in 2005 is Helen (d. 2007).
Step 2: Determine who Helen was married to. Helen was married to Sam Walton, the founder of Walmart and Sam's Club.
Step 3: Identify the current spouse of Helen. Christy Walton took her husband John's place after his death.
Step 4: Determine who John is. John is Christy Walton's husband.
Step 5: Combine the information to answer the question. Helen's husband after John died in 2005 is John, who is also Christy Walton's husband.
Therefore, the answer is: John.

--------------------
CoT-3: (*@$P(r_3|X) = 0.075$@*)
--------------------
Step 1: Identify the prominent Walton family member who died after John died in 2005. The prominent Walton family member who died after John died in 2005 is Helen Robson Kemper Walton.
Step 2: Determine the husband of Helen Robson Kemper Walton. Helen Robson Kemper Walton was the wife of Sam Walton.
Step 3: Combine the information to answer the question. Since John died in 2005, it is likely that the question is asking about the husband of Helen Robson Kemper Walton after John's death. Since Helen was married to Sam Walton, the correct answer is that there is no husband of Helen Robson Kemper Walton after John's death, as Sam Walton passed away before John.
Therefore, the answer is: There is no answer (as Helen Robson Kemper Walton was already married to Sam Walton and John had passed away).

--------------------
CoT-4: (*@$P(r_4|X) = 0.125$@*)
--------------------
Step 1: Identify the prominent Walton family member who died after John died in 2005. The prominent Walton family member who died after John died in 2005 is Helen Robson Kemper Walton.
Step 2: Determine who Helen's husband was. Helen's husband was Sam Walton, the founder of Walmart and Sam's Club.
Step 3: Since John died in 2005, it is likely that Helen's husband John is referring to John Walton, who died in 2005. This means Christy Walton took John's place after his death, as stated in the context.
Step 4: Therefore, the husband of the prominent Walton family member who died after John died in 2005 is John Walton.
Therefore, the answer is: John Walton.

--------------------
CoT-5: (*@$P(r_5|X) = 0.125$@*)
--------------------
Step 1: Identify the prominent Walton family member who died after John died in 2005. Helen Robson Kemper Walton died in 2007, after John Walton's death in 2005.
Step 2: Determine who Helen Robson Kemper Walton was married to. Helen Robson Kemper Walton was married to Sam Walton.
Step 3: Determine who took Sam Walton's place after his death. Sam Walton's wife Helen took her husband's place after his death.
Step 4: Identify who took John Walton's place after his death. Christy Walton took her husband John's place after his death.
Step 5: Combine the information to answer the question. Since Christy Walton took John's place after his death, it can be inferred that John was Christy's husband.
Step 6: Therefore, the answer is: John Walton.

--------------------
CoT-6: (*@$P(r_6|X) = 0.025$@*)
--------------------
Step 1: Identify the Walton family member who died after John died in 2005. The Walton family member who died after John died in 2005 is Helen Robson Kemper Walton.
Step 2: Determine who Helen Robson Kemper Walton was married to. According to the context, Helen was the wife of Sam Walton, the founder of Walmart and Sam's Club.
Step 3: Since John is mentioned as dying in 2005, it implies that Helen's husband is not John, but rather Sam Walton. However, the question asks about the husband of the prominent Walton family member who died after John died in 2005, which would be Helen.
Step 4: Since Helen is the only prominent Walton family member mentioned as dying after John died in 2005, it can be inferred that her husband is the one being asked about. However, the question is asking about the husband of the prominent Walton family member who died after John died in 2005, not Helen herself. This implies that the question is actually asking about the husband of Christy Walton, who took her husband John's place after his death.
Step 5: Therefore, the answer is: John Walton, who was Christy Walton's husband.

--------------------
CoT-7: (*@$P(r_7|X) = 0.05$@*)
--------------------
Step 1: Identify the Walton family member who died after John died in 2005. Helen Robson Kemper Walton died in 2007, after John's death in 2005.
Step 2: Determine who Helen's husband was. Helen's husband was Sam Walton, the co-founder of Walmart.
Step 3: However, the question asks for the husband of the prominent Walton family member who died after John died in 2005. Since Helen died in 2007, it cannot be her.
Step 4: The question does not provide the exact date of Christy Walton's husband John's death, but it mentions that Christy took her husband John's place after his death. Therefore, it can be inferred that Christy Walton's husband is the answer to the question.
Step 5: Combine the information to answer the question. The husband of the prominent Walton family member who died after John died in 2005 is John Walton's husband, which is Christy Walton's husband, also named John.
Therefore, the answer is: John.

--------------------
CoT-8: (*@$P(r_8|X) = 0.025$@*)
--------------------
Step 1: Identify the prominent Walton family member who died after John died in 2005. Helen Robson Kemper Walton died in 2007.
Step 2: Determine the relationship between Helen and John. Helen was the wife of John.
Step 3: Identify the question being asked. The question is asking for the husband of the prominent Walton family member who died after John died in 2005.
Step 4: Combine the information to answer the question. Since Helen died after John in 2005, John is the husband of the prominent Walton family member who died after John died in 2005.
Therefore, the answer is: John.

\end{lstlisting}
\textbf{Intervention results}
\begin{lstlisting}[breaklines=true, xleftmargin=0pt,breakindent=0pt, columns=fullflexible]
For the above eight chain-of-thoughts, we use the NWGM algorithm to perform causal intervention operations on them respectively. We can then compute the causal effect between the chain-of-thought and the answer.

(*@$P(A=John|do(r_1)) = 1/10 = 0.1$@*)
(*@$P(A=Sam Walton|do(r_1)) = 6/10 = 0.6$@*)
(*@$P(A=John Walton|do(r_1)) = 2/10 = 0.2$@*)

(*@$P(A=John|do(r_2)) = 3/10 = 0.3$@*)
(*@$P(A=Sam Walton|do(r_2)) = 4/10 = 0.4$@*)
(*@$P(A=John Walton|do(r_2)) = 1/10 = 0.1$@*)

(*@$P(A=John|do(r_3)) = 4/10 = 0.4$@*)
(*@$P(A=Sam Walton|do(r_3)) = 0/10 = 0.0$@*)
(*@$P(A=John Walton|do(r_3)) = 1/10 = 0.1$@*)

(*@$P(A=John|do(r_4)) = 0/10 = 0.0$@*)
(*@$P(A=Sam Walton|do(r_4)) = 2/10 = 0.2$@*)
(*@$P(A=John Walton|do(r_4)) = 8/10 = 0.8$@*)

(*@$P(A=John|do(r_5)) = 0/10 = 0.0$@*)
(*@$P(A=Sam Walton|do(r_5)) = 0/10 = 0.0$@*)
(*@$P(A=John Walton|do(r_5)) = 5/10 = 0.5$@*)

(*@$P(A=John|do(r_6)) = 0/10 = 0.0$@*)
(*@$P(A=Sam Walton|do(r_6)) = 0/10 = 0.0$@*)
(*@$P(A=John Walton|do(r_6)) = 8/10 = 0.8$@*)

(*@$P(A=John|do(r_7)) = 8/10 = 0.8$@*)
(*@$P(A=Sam Walton|do(r_7)) = 0/10 = 0.0$@*)
(*@$P(A=John Walton|do(r_7)) = 2/10 = 0.2$@*)

(*@$P(A=John|do(r_8)) = 6/10 = 0.6$@*)
(*@$P(A=Sam Walton|do(r_8)) = 0/10 = 0.0$@*)
(*@$P(A=John Walton|do(r_8)) = 1/10 = 0.1$@*)

\end{lstlisting}

\textbf{Final results}
\begin{lstlisting}[breaklines=true, xleftmargin=0pt,breakindent=0pt, columns=fullflexible]
The final answer is obtained by performing a weighted voting as follows:
(*@$P(A=John|do(X))$@*) = 0.325 * 0.1 + 0.25 * 0.3 + 0.075 * 0.4 + 0.125 * 0.0 + 0.125 * 0.0 + 0.025 * 0.0 + 0.05 * 0.8 + 0.025 * 0.6 = 0.1925

(*@$P(A=Sam Walton|do(X))$@*) = 0.325 * 0.6 + 0.25 * 0.4 + 0.075 * 0.0 + 0.125 * 0.2 + 0.125 * 0.0 + 0.025 * 0.0 + 0.05 * 0.0 + 0.025 * 0.0 = 0.32

(*@$P(A=John Walton|do(X))$@*) = 0.325 * 0.2 + 0.25 * 0.1 + 0.075 * 0.1 + 0.125 * 0.8 + 0.125 * 0.5 + 0.025 * 0.8 + 0.05 * 0.2 + 0.025 * 0.1 = 0.2925
Finally, we chose the answer with the largest weight as the final answer.
Therefore, the final answer obtained according to the Causal Prompting method is (*@\textcolor{c2}{Sam Walton}@*).
\end{lstlisting}
\end{tcolorbox}

\clearpage
\newpage
\section{Prompt Templates}
\label{sec:prompt}

In this section, we introduce the prompt templates of Chain-of-thought prompting (detailed in Section 3.1), CoT Improvement based on NWGM approximation (detailed in Section 3.2), Samples generation for Contrastive Learning (detailed in Section 3.4) and Demonstration Construction (detailed in Appendix~\ref{sec:settings}), respectively.
The \textcolor{c1}{blue} texts in prompts are required for LLM completion.

\subsection{Chain-of-thought prompting}
\label{sec:prompt_cot}

\phantomsection
\label{fig:template_multihop_qa_cot}
\begin{tcolorbox}[colback=white!95!gray,colframe=gray!50!black,rounded corners,label={template-second-select-roles}, title={CoT Prompt template of Multi-hop Question Answering task.}]
\textbf{Instruction}
\begin{lstlisting}[breaklines=true, xleftmargin=0pt, breakindent=0pt, columns=fullflexible]
You are a helpful assistant to perform Multi-hop Question Answering. Based on the context, answer the question step by step and provide the final answer in the end.
\end{lstlisting}
\textbf{Demonstration}
\begin{lstlisting}[breaklines=true, xleftmargin=0pt, breakindent=0pt, columns=fullflexible]
Q: 
The context is: [paragraphs]
The question is: [question]
Let us think step by step.
A:
Sure! Let us think step by step. [cot]
Therefore, the final answer is: [answer]
\end{lstlisting}
\textbf{Test example:}
\begin{lstlisting}[breaklines=true, xleftmargin=0pt,breakindent=0pt, columns=fullflexible]
Q: 
The context is: [paragraphs]
The question is: [question]
Let us think step by step.
A:
(*@\textcolor{c1}{
Sure! Let us think step by step. [cot]\\
Therefore, the final answer is: [answer]
}@*)
\end{lstlisting}
\end{tcolorbox}

\phantomsection
\label{fig:template_GSM8K_cot}
\begin{tcolorbox}[colback=white!95!gray,colframe=gray!50!black,rounded corners,label={template-second-select-roles}, title={CoT Prompt template of GSM8K dataset.}]
\textbf{Instruction}
\begin{lstlisting}[breaklines=true, xleftmargin=0pt, breakindent=0pt, columns=fullflexible]
You are a helpful assistant to perform Mathematical reasoning. Answer the question step by step and provide the final answer in the end.
\end{lstlisting}
\textbf{Demonstration}
\begin{lstlisting}[breaklines=true, xleftmargin=0pt, breakindent=0pt, columns=fullflexible]
Q:
The question is: [question]
Let us think step by step.
A:
Sure! Let us think step by step. [cot]
Therefore, the final answer is: [answer]
\end{lstlisting}
\textbf{Test example:}
\begin{lstlisting}[breaklines=true, xleftmargin=0pt,breakindent=0pt, columns=fullflexible]
Q:
The question is: [question]
Let us think step by step.
A:
(*@\textcolor{c1}{
Sure! Let us think step by step. [cot]\\
Therefore, the final answer is: [answer]
}@*)
\end{lstlisting}
\end{tcolorbox}

\phantomsection
\label{fig:template_MATH_cot}
\begin{tcolorbox}[colback=white!95!gray,colframe=gray!50!black,rounded corners,label={template-second-select-roles}, title={CoT Prompt template of MATH dataset.}]
\textbf{Instruction}
\begin{lstlisting}[breaklines=true, xleftmargin=0pt, breakindent=0pt, columns=fullflexible]
You are a helpful assistant to perform Mathematical reasoning. Answer the question step by step and provide the final answer in the end. Presented in Latex format in text mode. Your answer should be inside \boxed{}, such as \boxed{answer}.
\end{lstlisting}
\textbf{Demonstration}
\begin{lstlisting}[breaklines=true, xleftmargin=0pt, breakindent=0pt, columns=fullflexible]
Q:
The question is: [problem]
Let us think step by step.
A:
Sure! Let us think step by step. [cot]
Therefore, the final answer is: [answer]
\end{lstlisting}
\textbf{Test example:}
\begin{lstlisting}[breaklines=true, xleftmargin=0pt,breakindent=0pt, columns=fullflexible]
Q:
The question is: [problem]
Let us think step by step.
A:
(*@\textcolor{c1}{
Sure! Let us think step by step. [cot]\\
Therefore, the final answer is: [answer]
}@*)
\end{lstlisting}
\end{tcolorbox}

\phantomsection
\label{fig:template_ABSA_cot}
\begin{tcolorbox}[colback=white!95!gray,colframe=gray!50!black,rounded corners,label={template-second-select-roles}, title={CoT Prompt template of Aspect-based Sentiment Analysis (ABSA) task.}]
\textbf{Instruction}
\begin{lstlisting}[breaklines=true, xleftmargin=0pt, breakindent=0pt, columns=fullflexible]
You are a helpful assistant to perform sentiment classification. Please detect the sentiment polarity towards the target given the sentence. The sentiment polarities include positive, negative and neutral. Please focus on sentiment of the target itself. Detect the sentiment polarity step by step and provide the final answer in the end.
\end{lstlisting}
\textbf{Demonstration}
\begin{lstlisting}[breaklines=true, xleftmargin=0pt, breakindent=0pt, columns=fullflexible]
Q: 
The sentence is: [text] 
The target is: [target].
Let us think step by step.
A:
Sure! Let us think step by step. [cot]
Therefore, the final answer is: [answer]
\end{lstlisting}
\textbf{Test example:}
\begin{lstlisting}[breaklines=true, xleftmargin=0pt,breakindent=0pt, columns=fullflexible]
Q: 
The sentence is: [text] 
The target is: [target].
Let us think step by step.
A:
(*@\textcolor{c1}{
Sure! Let us think step by step. [cot]\\
Therefore, the final answer is: [answer]
}@*)
\end{lstlisting}
\end{tcolorbox}

\phantomsection
\label{fig:template_NLI_cot}
\begin{tcolorbox}[colback=white!95!gray,colframe=gray!50!black,rounded corners,label={template-second-select-roles}, title={CoT Prompt template of Natural Language Inference (NLI) task.}]
\textbf{Instruction}
\begin{lstlisting}[breaklines=true, xleftmargin=0pt, breakindent=0pt, columns=fullflexible]
You are a helpful assistant to perform Natural language inference. Natural language inference is the task of determining whether a "hypothesis" is true (entailment), false (contradiction), or undetermined (neutral) given a "premise". Answer in a consistent style. Please write the reasoning process before giving the answer. Please provide your answer in the last sentence of your response. Your answer should be entailment, contradiction or neutral.
\end{lstlisting}
\textbf{Demonstration}
\begin{lstlisting}[breaklines=true, xleftmargin=0pt, breakindent=0pt, columns=fullflexible]
Q: 
The premise is: [premise] 
The hypothesis is: [hypothesis] 
Let us think step by step.
A:
Sure! Let us think step by step. [cot]
Therefore, the final answer is: [answer]
\end{lstlisting}
\textbf{Test example:}
\begin{lstlisting}[breaklines=true, xleftmargin=0pt,breakindent=0pt, columns=fullflexible]
Q: 
The premise is: [premise] 
The hypothesis is: [hypothesis] 
Let us think step by step.
A:
(*@\textcolor{c1}{
Sure! Let us think step by step. [cot]\\
Therefore, the final answer is: [answer]
}@*)
\end{lstlisting}
\end{tcolorbox}

\phantomsection
\label{fig:template_fever_cot}
\begin{tcolorbox}[colback=white!95!gray,colframe=gray!50!black,rounded corners,label={template-second-select-roles}, title={CoT Prompt template of Fact Verification (FV) task.}]
\textbf{Instruction}
\begin{lstlisting}[breaklines=true, xleftmargin=0pt, breakindent=0pt, columns=fullflexible]
You are a helpful assistant to perform fact verification. Please check the veracity of the claim according to the evidence, including SUPPORTS and REFUTES. Answer in a consistent style. Please write the reasoning process before giving the answer. Please provide your answer in the last sentence of your response. Your answer should be either SUPPORTS or REFUTES.
\end{lstlisting}
\textbf{Demonstration}
\begin{lstlisting}[breaklines=true, xleftmargin=0pt, breakindent=0pt, columns=fullflexible]
Q: 
The claim is: [question] 
The evidence is: [context]
Let us think step by step.
A:
Sure! Let us think step by step. [cot]
Therefore, the final answer is: [answer]
\end{lstlisting}
\textbf{Test example:}
\begin{lstlisting}[breaklines=true, xleftmargin=0pt,breakindent=0pt, columns=fullflexible]
Q: 
The claim is: [question] 
The evidence is: [context]
Let us think step by step.
A:
(*@\textcolor{c1}{
Sure! Let us think step by step. [cot]\\
Therefore, the final answer is: [answer]
}@*)
\end{lstlisting}
\end{tcolorbox}

\subsection{CoT Improvement based on NWGM approximation}
\label{sec:prompt_causal_prompt}
We show the prompt templates of CoT Improvement based on NWGM approximation for the seven datasets listed below, including HotpotQA, MuSiQue, GSM8K, MATH, ABSA, NLI and FV. Among them, HotpotQA and MuSiQue datasets share the same template.

\phantomsection
\label{fig:template_multihop_qa}
\begin{tcolorbox}[colback=white!95!gray,colframe=gray!50!black,rounded corners,label={template-second-select-roles}, title={Prompt template of Multi-hop Question Answering task.}]
\textbf{Instruction}
\begin{lstlisting}[breaklines=true, xleftmargin=0pt, breakindent=0pt, columns=fullflexible]
You are a helpful assistant to perform Multi-hop Question Answering. Based on the context, answer the question step by step and provide the final answer in the end. I will provide a reasoning process, and please improve the reasoning process and make sure you get the correct answer. Give your final answer using the "The answer is:" format.
\end{lstlisting}
\textbf{Demonstration}
\begin{lstlisting}[breaklines=true, xleftmargin=0pt, breakindent=0pt, columns=fullflexible]
Q: 
The context is: [paragraphs]
The question is: [question]
Let us think step by step.
The provided reasoning process is: [wrong_cot]
A:
The improved reasoning process is: [correct_cot]
Therefore, the correct answer is: [answer]
\end{lstlisting}
\textbf{Test example:}
\begin{lstlisting}[breaklines=true, xleftmargin=0pt,breakindent=0pt, columns=fullflexible]
Q: 
The context is: [paragraphs]
The question is: [question]
Let us think step by step.
The provided reasoning process is: [wrong_cot]
A:
(*@\textcolor{c1}{
The improved reasoning process is: [improved\_cot]\\
Therefore, the correct answer is: [answer]
}@*)
\end{lstlisting}
\end{tcolorbox}

\phantomsection
\label{fig:template_GSM8K}
\begin{tcolorbox}[colback=white!95!gray,colframe=gray!50!black,rounded corners,label={template-second-select-roles}, title={Prompt template of GSM8K dataset.}]
\textbf{Instruction}
\begin{lstlisting}[breaklines=true, xleftmargin=0pt, breakindent=0pt, columns=fullflexible]
You are a helpful assistant to perform Mathematical reasoning. Answer the question step by step
and provide the final answer in the end. I will provide a reasoning process, and please improve the
reasoning process and make sure you get the correct answer.
\end{lstlisting}
\textbf{Demonstration}
\begin{lstlisting}[breaklines=true, xleftmargin=0pt, breakindent=0pt, columns=fullflexible]
Q:
The question is: [question]
Let us think step by step.
The provided reasoning process is: [wrong_cot]
A:
The improved reasoning process is: [correct_cot]
Therefore, the correct answer is: [answer]
\end{lstlisting}
\textbf{Test example:}
\begin{lstlisting}[breaklines=true, xleftmargin=0pt,breakindent=0pt, columns=fullflexible]
Q:
The question is: [question]
Let us think step by step.
The provided reasoning process is: [wrong_cot]
A:
(*@\textcolor{c1}{
The improved reasoning process is: [improved\_cot]\\
Therefore, the correct answer is: [answer]
}@*)
\end{lstlisting}
\end{tcolorbox}

\phantomsection
\label{fig:template_MATH}
\begin{tcolorbox}[colback=white!95!gray,colframe=gray!50!black,rounded corners,label={template-second-select-roles}, title={Prompt template of MATH dataset.}]
\textbf{Instruction}
\begin{lstlisting}[breaklines=true, xleftmargin=0pt, breakindent=0pt, columns=fullflexible]
You are a helpful assistant to perform Mathematical reasoning. Answer the question step by step and provide the final answer in the end. I will provide a reasoning process, and please improve the reasoning process and make sure you get the correct answer. Presented in Latex format in text mode. Your answer should be inside \boxed{}, such as \boxed{answer}.
\end{lstlisting}
\textbf{Demonstration}
\begin{lstlisting}[breaklines=true, xleftmargin=0pt, breakindent=0pt, columns=fullflexible]
Q: 
The question is: [problem]
Let us think step by step.
The provided reasoning process is: [wrong_cot]
A:
The improved reasoning process is: [correct_cot]
Therefore, the correct answer is: [answer]
\end{lstlisting}
\textbf{Test example:}
\begin{lstlisting}[breaklines=true, xleftmargin=0pt,breakindent=0pt, columns=fullflexible]
Q: 
The question is: [problem]
Let us think step by step.
The provided reasoning process is: [wrong_cot]
A:
(*@\textcolor{c1}{
The improved reasoning process is: [improved\_cot]\\
Therefore, the correct answer is: [answer]
}@*)
\end{lstlisting}
\end{tcolorbox}

\phantomsection
\label{fig:template_ABSA}
\begin{tcolorbox}[colback=white!95!gray,colframe=gray!50!black,rounded corners,label={template-second-select-roles}, title={Prompt template of Aspect-based Sentiment Analysis (ABSA) task.}]
\textbf{Instruction}
\begin{lstlisting}[breaklines=true, xleftmargin=0pt, breakindent=0pt, columns=fullflexible]
You are a helpful assistant to perform sentiment classification. Please detect the sentiment polarity towards the target given the sentence. The sentiment polarities include positive, negative and neutral. Please focus on sentiment of the target itself. Detect the sentiment polarity step by step and provide the final answer in the end. I will provide a reasoning process, and please improve the reasoning process and make sure you get the correct answer.
\end{lstlisting}
\textbf{Demonstration}
\begin{lstlisting}[breaklines=true, xleftmargin=0pt, breakindent=0pt, columns=fullflexible]
Q: 
The sentence is: [text] 
The target is: [target].
Let us think step by step.
The provided reasoning process is: [wrong_cot]
A:
The improved reasoning process is: [correct_cot]
Therefore, the correct answer is: [answer]
\end{lstlisting}
\textbf{Test example:}
\begin{lstlisting}[breaklines=true, xleftmargin=0pt,breakindent=0pt, columns=fullflexible]
Q: 
The sentence is: [text] 
The target is: [target].
Let us think step by step.
The provided reasoning process is: [wrong_cot]
A:
(*@\textcolor{c1}{
The improved reasoning process is: [improved\_cot]\\
Therefore, the correct answer is: [answer]
}@*)
\end{lstlisting}
\end{tcolorbox}

\phantomsection
\label{fig:template_NLI}
\begin{tcolorbox}[colback=white!95!gray,colframe=gray!50!black,rounded corners,label={template-second-select-roles}, title={Prompt template of Natural Language Inference (NLI) task.}]
\textbf{Instruction}
\begin{lstlisting}[breaklines=true, xleftmargin=0pt, breakindent=0pt, columns=fullflexible]
You are a helpful assistant to perform Natural language inference. Natural language inference is the task of determining whether a "hypothesis" is true (entailment), false (contradiction), or undetermined (neutral) given a "premise". Answer in a consistent style. Please write the reasoning process before giving the answer. Please provide your answer in the last sentence of your response. Your answer should be entailment, contradiction or neutral. I will provide a reasoning process, and please improve the reasoning process and make sure you get the correct answer.
\end{lstlisting}
\textbf{Demonstration}
\begin{lstlisting}[breaklines=true, xleftmargin=0pt, breakindent=0pt, columns=fullflexible]
Q: 
The premise is: [premise] 
The hypothesis is: [hypothesis] 
Let us think step by step.
The provided reasoning process is: [wrong_cot]
A:
The improved reasoning process is: [correct_cot]
Therefore, the correct answer is: [answer]
\end{lstlisting}
\textbf{Test example:}
\begin{lstlisting}[breaklines=true, xleftmargin=0pt,breakindent=0pt, columns=fullflexible]
Q: 
The premise is: [premise] 
The hypothesis is: [hypothesis] 
Let us think step by step.
The provided reasoning process is: [wrong_cot]
A:
(*@\textcolor{c1}{
The improved reasoning process is: [improved\_cot]\\
Therefore, the correct answer is: [answer]
}@*)
\end{lstlisting}
\end{tcolorbox}

\phantomsection
\label{fig:template_fever}
\begin{tcolorbox}[colback=white!95!gray,colframe=gray!50!black,rounded corners,label={template-second-select-roles}, title={Prompt template of Fact Verification (FV) task.}]
\textbf{Instruction}
\begin{lstlisting}[breaklines=true, xleftmargin=0pt, breakindent=0pt, columns=fullflexible]
You are a helpful assistant to perform fact verification. Please check the veracity of the claim according to the evidence, including SUPPORTS and REFUTES. Answer in a consistent style. Please write the reasoning process before giving the answer. Please provide your answer in the last sentence of your response. Your answer should be either SUPPORTS or REFUTES. I will provide a reasoning process, and please improve the reasoning process and make sure you get the correct answer.
\end{lstlisting}
\textbf{Demonstration}
\begin{lstlisting}[breaklines=true, xleftmargin=0pt, breakindent=0pt, columns=fullflexible]
Q: 
The claim is: [question] 
The evidence is: [context]
Let us think step by step.
The provided reasoning process is: [wrong_cot]
A:
The improved reasoning process is: [correct_cot]
Therefore, the correct answer is: [answer]
\end{lstlisting}
\textbf{Test example:}
\begin{lstlisting}[breaklines=true, xleftmargin=0pt,breakindent=0pt, columns=fullflexible]
Q: 
The claim is: [question] 
The evidence is: [context]
Let us think step by step.
The provided reasoning process is: [wrong_cot]
A:
(*@\textcolor{c1}{
The improved reasoning process is: [improved\_cot]\\
Therefore, the correct answer is: [answer]
}@*)
\end{lstlisting}
\end{tcolorbox}

\subsection{Samples generation for Contrastive Learning}
\label{sec:prompt_contrastive_learning}

\phantomsection
\label{fig:template_cl}
\begin{tcolorbox}[colback=white!95!gray,colframe=gray!50!black,rounded corners,label={template-second-select-roles}, title={Prompt template of samples generation for Contrastive Learning.}]
\textbf{Instruction}
\begin{lstlisting}[breaklines=true, xleftmargin=0pt, breakindent=0pt, columns=fullflexible]
You are an expert in data augmentation. Please generate similar sentences based on the sentences I provided. Don't generate extraneous content.
\end{lstlisting}
\textbf{Test example:}
\begin{lstlisting}[breaklines=true, xleftmargin=0pt,breakindent=0pt, columns=fullflexible]
Q: 
Provided sentences: [anchor_sentences]
A:
(*@\textcolor{c1}{
Positive sentences: [positive\_sentences]
}@*)
\end{lstlisting}
\end{tcolorbox}

\subsection{Demonstration Construction}
\label{sec:prompt_demo_gen}

\phantomsection
\label{fig:template_multihop_qa_demo_gen}
\begin{tcolorbox}[colback=white!95!gray,colframe=gray!50!black,rounded corners,label={template-second-select-roles}, title={Prompt template of demos generation for Multi-hop Question Answering task.}]
\textbf{Instruction}
\begin{lstlisting}[breaklines=true, xleftmargin=0pt, breakindent=0pt, columns=fullflexible]
You are a helpful assistant to perform Multi-hop Question Answering. Based on the context, answer the question step by step and provide the final answer in the end. I will provide the correct answer and ask you to write your thought process based on the answer.
\end{lstlisting}
\textbf{Demonstration}
\begin{lstlisting}[breaklines=true, xleftmargin=0pt, breakindent=0pt, columns=fullflexible]
Q: 
The context is: [paragraphs]
The question is: [question]
The correct answer is: [answer]
Let us think step by step.
A:
The correct reasoning process is: [cot]
\end{lstlisting}
\textbf{Test example:}
\begin{lstlisting}[breaklines=true, xleftmargin=0pt,breakindent=0pt, columns=fullflexible]
Q: 
The context is: [paragraphs]
The question is: [question]
The correct answer is: [answer]
Let us think step by step.
A:
(*@\textcolor{c1}{
The correct reasoning process is: [cot]
}@*)
\end{lstlisting}
\end{tcolorbox}

\phantomsection
\label{fig:template_ABSA_demo_gen}
\begin{tcolorbox}[colback=white!95!gray,colframe=gray!50!black,rounded corners,label={template-second-select-roles}, title={Prompt template of demos generation for Aspect-based Sentiment Analysis (ABSA) task.}]
\textbf{Instruction}
\begin{lstlisting}[breaklines=true, xleftmargin=0pt, breakindent=0pt, columns=fullflexible]
You are a helpful assistant to perform sentiment classification. Please detect the sentiment polarity towards the target given the sentence. The sentiment polarities include positive, negative and neutral. Please focus on sentiment of the target itself. Detect the sentiment polarity step by step and provide the final answer in the end. I will provide the correct answer and ask you to write your thought process based on the answer.
\end{lstlisting}
\textbf{Demonstration}
\begin{lstlisting}[breaklines=true, xleftmargin=0pt, breakindent=0pt, columns=fullflexible]
Q: 
The sentence is: [text] 
The target is: [target]. 
The correct answer is: [answer]
Let us think step by step.
A:
The correct reasoning process is: [cot]
\end{lstlisting}
\textbf{Test example:}
\begin{lstlisting}[breaklines=true, xleftmargin=0pt,breakindent=0pt, columns=fullflexible]
Q: 
The sentence is: [text] 
The target is: [target]. 
The correct answer is: [answer]
Let us think step by step.
A:
(*@\textcolor{c1}{
The correct reasoning process is: [cot]
}@*)
\end{lstlisting}
\end{tcolorbox}

\phantomsection
\label{fig:template_NLI_demo_gen}
\begin{tcolorbox}[colback=white!95!gray,colframe=gray!50!black,rounded corners,label={template-second-select-roles}, title={Prompt template of demos generation for Natural Language Inference (NLI) task.}]
\textbf{Instruction}
\begin{lstlisting}[breaklines=true, xleftmargin=0pt, breakindent=0pt, columns=fullflexible]
You are a helpful assistant to perform Natural language inference. Natural language inference is the task of determining whether a "hypothesis" is true (entailment), false (contradiction), or undetermined (neutral) given a "premise". Answer in a consistent style. Please write the reasoning process before giving the answer. Please provide your answer in the last sentence of your response. Your answer should be entailment, contradiction or neutral. I will provide the correct answer and ask you to write your thought process based on the answer.
\end{lstlisting}
\textbf{Demonstration}
\begin{lstlisting}[breaklines=true, xleftmargin=0pt, breakindent=0pt, columns=fullflexible]
Q: 
The premise is: [premise] 
The hypothesis is: [hypothesis] 
The correct answer is: [answer]
Let us think step by step.
A:
The correct reasoning process is: [cot]
\end{lstlisting}
\textbf{Test example:}
\begin{lstlisting}[breaklines=true, xleftmargin=0pt,breakindent=0pt, columns=fullflexible]
Q: 
The premise is: [premise] 
The hypothesis is: [hypothesis] 
The correct answer is: [answer]
Let us think step by step.
A:
(*@\textcolor{c1}{
The correct reasoning process is: [cot]
}@*)
\end{lstlisting}
\end{tcolorbox}

\phantomsection
\label{fig:template_fever_demo_gen}
\begin{tcolorbox}[colback=white!95!gray,colframe=gray!50!black,rounded corners,label={template-second-select-roles}, title={Prompt template of demos generation for Fact Verification (FV) task.}]
\textbf{Instruction}
\begin{lstlisting}[breaklines=true, xleftmargin=0pt, breakindent=0pt, columns=fullflexible]
You are a helpful assistant to perform fact verification. Please check the veracity of the claim according to the evidence, including SUPPORTS and REFUTES. Answer in a consistent style. Please write the reasoning process before giving the answer. Please provide your answer in the last sentence of your response. Your answer should be either SUPPORTS or REFUTES. I will provide the correct answer and ask you to write your thought process based on the answer.
\end{lstlisting}
\textbf{Demonstration}
\begin{lstlisting}[breaklines=true, xleftmargin=0pt, breakindent=0pt, columns=fullflexible]
Q: 
The claim is: [question] 
The evidence is: [context]
The correct answer is: [answer]
Let us think step by step.
A:
The correct reasoning process is: [cot]
\end{lstlisting}
\textbf{Test example:}
\begin{lstlisting}[breaklines=true, xleftmargin=0pt,breakindent=0pt, columns=fullflexible]
Q: 
The claim is: [question] 
The evidence is: [context]
The correct answer is: [answer]
Let us think step by step.
A:
(*@\textcolor{c1}{
The correct reasoning process is: [cot]
}@*)
\end{lstlisting}
\end{tcolorbox}
